\pdfoutput=1

\documentclass[11pt]{article}

\usepackage[]{ACL2023}

\usepackage{times}
\usepackage{latexsym}
\usepackage{graphicx} 
\usepackage{amsmath}
\usepackage{amssymb}
\usepackage{subcaption}

\usepackage[T1]{fontenc}

\usepackage[utf8]{inputenc}

\usepackage{microtype}

\usepackage{inconsolata}

\newcommand{\redtxt}[1]{\textcolor{red}{#1}}
\newcommand{\ourmethod}{TwO}

\title{Combo of Thinking  and Observing  for  Outside-Knowledge VQA }

\author{Qingyi Si$^{1,2}$,\ Yuchen Mo$^{3}$,\ Zheng Lin$^{1,2}$\thanks{\ \ Corresponding author: Zheng Lin.},\ Huishan Ji$^{1,2}$,\ Weiping Wang$^1$  \\ 
$^1$Institute of Information Engineering, Chinese Academy of Sciences, Beijing, China \\
$^2$School of Cyber Security, University of Chinese Academy of Sciences, Beijing, China \\
$^3$ByteDance AI Lab, Beijing, China  \\
  \{siqingyi,linzheng,jihuishan,wangweiping\}@iie.ac.cn \\
}

\begin{document}
\maketitle
\begin{abstract}

Outside-knowledge visual question answering is a challenging task that requires both the acquisition and the use of open-ended real-world knowledge. 
Some existing solutions draw external knowledge into the cross-modality space which overlooks the much vaster textual knowledge in natural-language space, while others transform the image into a text that further fuses with the textual knowledge into the natural-language space and completely abandons the use of visual features. 
In this paper, we are inspired to constrain the cross-modality space into the same space of natural-language space which makes  
the visual features preserved directly, and the model still benefits from the vast knowledge in natural-language space. 
To this end, we propose a novel framework consisting of a multimodal encoder, a textual encoder and an answer decoder. 
Such structure allows us to introduce more types of knowledge including explicit and implicit multimodal and textual knowledge.
Extensive experiments validate the superiority of the proposed method which outperforms the state-of-the-art by 6.17\% accuracy.
We also conduct comprehensive ablations of each component, and systematically study the roles of varying types of knowledge. Codes and knowledge data can be found at \url{https://github.com/PhoebusSi/Thinking-while-Observing}.\footnote{ Joint work with ByteDance AI Lab. }  





\end{abstract}

\section{Introduction}

\begin{figure}[t]
\centering
\includegraphics[width=0.95\columnwidth]{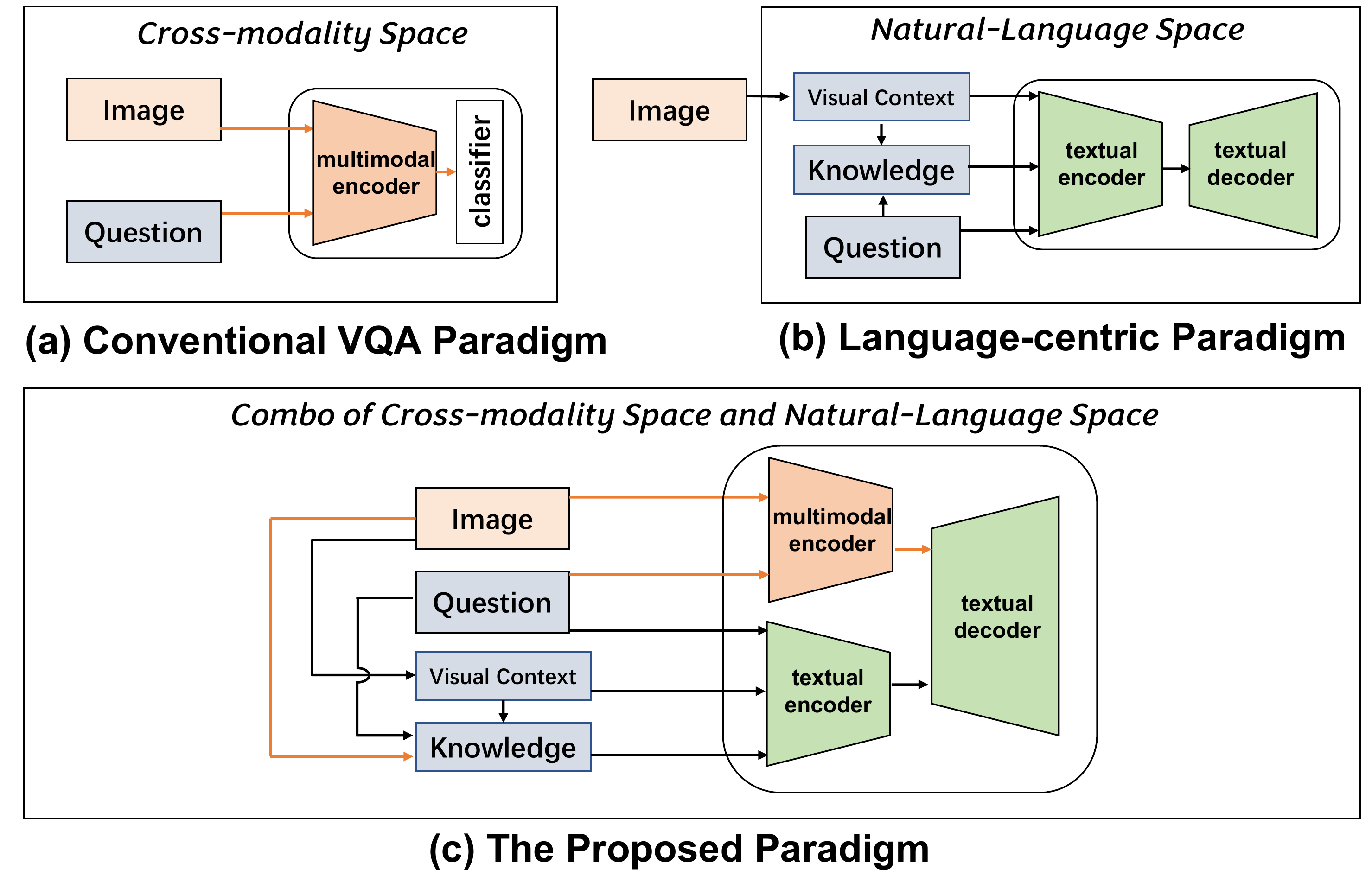} 
\caption{Comparison with previous paradigms. Orange lines indicate processes involving cross-modality space. (a) The conventional VQA paradigm fuses image and question text into the cross-modality space, and then predicts answers in a close-set classification manner. (b) Language-centric paradigm applies captioning and tagging tools to describe the visual context, and abandons the visual features to convert the VQA task into an open-ended generative QA task. (c) The proposed paradigm intends to constrain the cross-modality space into the same space as natural-language space so that models can directly decode both text and multimodal embeddings. }
\label{paradigm}
\end{figure}


Conventional visual question answering (VQA) \citep{antol2015vqa} tasks require models to answer questions based on image content. Such tasks have been thoroughly studied \citep{guo2021bilinear,jiang2020defense,li2020unimo} on conventional VQA datasets VQAv2~\citep{goyal2017making}. 
However, real-world questions often rely on a certain amount of knowledge beyond images. 
Therefore, Knowledge Base Question Answering (KB-VQA) tasks
\citep{cao2021knowledge,wang2015explicit,wang2017fvqa,shah2019kvqa,lu2018r} always require models to answer questions by referring to the corresponding knowledge facts in a specific pre-defined knowledge base. 
Yet any pre-defined knowledge base is far from covering real-world knowledge. 
Recently, the outside-knowledge visual question answering (OK-VQA) task has been proposed \citep{marino2019ok} and provides the most open VQA setting. That is, any knowledge resource can be used to answer its challenging and diverse questions. 

Most previous work \citep{ding2022mukea,garderes2020conceptbert,Marino_2021_CVPR} on OK-VQA follows conventional VQA paradigm (as shown in Figure \ref{paradigm} (a)) based on visual-language pre-trained (VLP) models, and injects knowledge into the same cross-modality space afterward. 
However, knowledge in cross-modality space is much less than that in natural-language space  \citeauthor{gao2022transform}. 
This paradigm excels at visual understanding, but refers to little knowledge, like a human who focuses on \textit{observing} but does not \textit{think} enough.

To take the advantage of the vast knowledge in natural-language space, 
state-of-the-art methods \citep{gao2022transform,yang2022empirical,gui2021kat} on OK-VQA follow language-centric paradigm (as shown in Figure \ref{paradigm} (b)) based on pre-trained language models (PLMs).
However, although more knowledge can be introduced, the paradigm is counter-intuitive because 
many visual details are lost when converting an image into text. Therefore, it is like a human who starts \textit{thinking} after brief \textit{observing}. 

For a human, a feasible solution to OK-VQA is combo \textit{\textbf{T}hinking \textbf{w}hile \textbf{O}bserving}. 
To this end, we propose \textbf{\ourmethod}, which is a framework consisting of a multimodal encoder, a textual encoder and an answer decoder. As shown in Figure \ref{paradigm}(c), the multimodal encoder directly encodes the visual features and acts as the role of \textit{observer}, while the textual encoder encodes a range of knowledge resources and acts as the role of \textit{thinker}. 
Finally, the answer decoder decodes the latent embeddings from both encoders 
to generate the final answer. In addition, a pre-training stage is added to help constrain the output of both encoders to the same latent space. 

\begin{figure}[t]
\centering
\includegraphics[width=0.9\columnwidth]{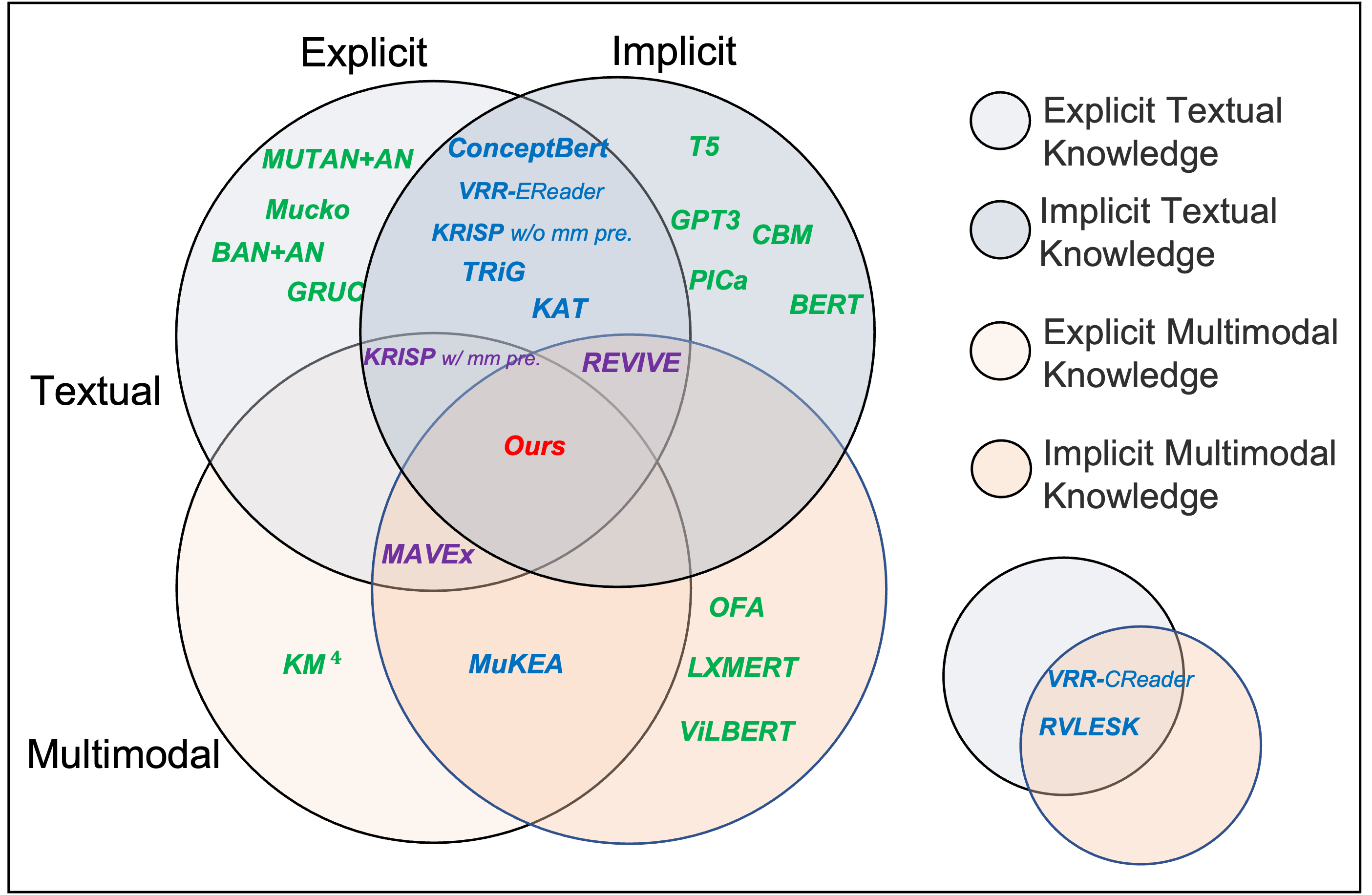} 
\caption{Taxonomy of OK-VQA methods by knowledge types. Green, purple, blue and red fonts represent the introduction of one, two, three and four types of knowledge. No existing work introduces four types of knowledge in a unified framework, but ours. }
\label{motivation}
\end{figure}

Previous methods \citep{gui2021kat, gao2022transform,wu2022multi} have thoroughly studied explicit textual knowledge such as Wikipedia, as well as implicit textual knowledge in GPT-3 \citep{brown2020language}. However, the discussion of multimodal knowledge, which further utilizes visual features, is still in its infancy in OK-VQA. 
In this paper, we accumulate explicit multimodal knowledge during pre-training on VQAv2 \citep{ding2022mukea}. Besides, inspired by prompting GPT-3 \citep{yang2022empirical} for implicit textual knowledge, 
we use prompt to bring in implicit multimodal knowledge stored in the unifying VLP model OFA \citep{wang2022ofa}.
Moreover, we refine a taxonomy of existing methods by knowledge (refer to Figure \ref{motivation}) 
where our method is the first to bring in all types of knowledge.

To summarize, our contributions are as follows:

(1) We propose a simple and effective paradigm that combines the advantages of both conventional VQA and language-centric paradigms. 

(2) Our method can deal with more comprehensive types of knowledge, and is the first to bring in implicit multimodal knowledge through a prompt-learning fashion. In addition, 
we empirically analyze the roles of different types of knowledge. 

(3) Experimental results show the effectiveness of our method, which establishes a new SoTA accuracy on OK-VQA with a 6.17\% gain.











\section{Background}

\subsection{Outside-Knowledge Visual Question Answering (OK-VQA)}

In addition to dividing existing methods according to latent space, namely multimodal-space methods \citep{ding2022mukea,garderes2020conceptbert,zhu2020mucko,yu2020cross, zheng2021km4,Marino_2021_CVPR} and textual-space methods \citep{yang2022empirical,gui2021kat,gao2022transform}, existing methods can also be roughly categorized into two lines by whether GPT-3 is used.  
Most of the GPT-3 based methods \citep{gui2021kat,lin2022revive} outperform non-GPT ones 
by large margins, since huge-parameter-capacity GPT-3 can store abundant implicit textual knowledge. The vast implicit  knowledge in GPT-3 can be easily retrieved in a prompt manner. 
For example, Pica \citep{yang2022empirical} uses text prompts of in-context examples to query GPT-3 for answers directly. 
However, most existing methods for OK-VQA are non-GPT-3 based, which do not directly compare with GPT-3 based methods for a fair comparison. For completeness, we explore our model performance with and without GPT-3, respectively. 


Previous work has generally improved model performance in OK-VQA in two ways: one is to introduce more knowledge sources (see Figure \ref{knows}), and the other is to optimize the model paradigm (see Figure \ref{paradigm}). 
For example, MAVEx \citep{wu2022multi} follows the former way and introduces more knowledge sources such as Wikipedia, ConceptNet \citep{speer2017conceptnet} and Google images to boost model performance; VRR-EReader \citep{luo2021weakly} follows the latter way and replaces the classifier with an extraction reader to solve the generalization problem of classification manner. 
Our method goes further in both directions: On the one hand, we explore more comprehensive types of knowledge. 
On the other hand, we refine the paradigm to make the visual features retained, and the model still benefits from natural language space. 
We list the relationship between our method and previous work in Appendix \ref{morerelation}.




\subsection{Taxonomy of OK-VQA Methods by Knowledge Types}
With an in-depth look at the types of knowledge involved in each existing method, we propose a complete taxonomy of OK-VQA methods shown in Figure \ref{motivation}. 
We divide all knowledge into four types: explicit textual knowledge, explicit multimodal knowledge, implicit textual knowledge, and implicit multimodal knowledge. 

 From Figure \ref{motivation}, we find that (1) most GPT-3 based methods \citep{yang2022empirical,gui2021kat} appear in the two circles of "Textual" because they adopt the language-centric paradigm. (2) There are few methods to use explicit multimodal knowledge, which is more challenging to introduce into models than explicit textual knowledge. Among them, \citeauthor{Marino_2021_CVPR,ding2022mukea} propose accumulating this knowledge through pre-training while \citeauthor{wu2022multi} use Google Image to provide similar images. (3) Recent work is usually distributed in the two circles of "Implicit". This shows that VLP models or PLMs have become one of the vital components of the model for OK-VQA. Appendix \ref{moreVLP} and \ref{moreLLM} show more related work about VLP models and PLMs.  

\begin{figure*}[t]
\centering
\includegraphics[width=0.98\textwidth]{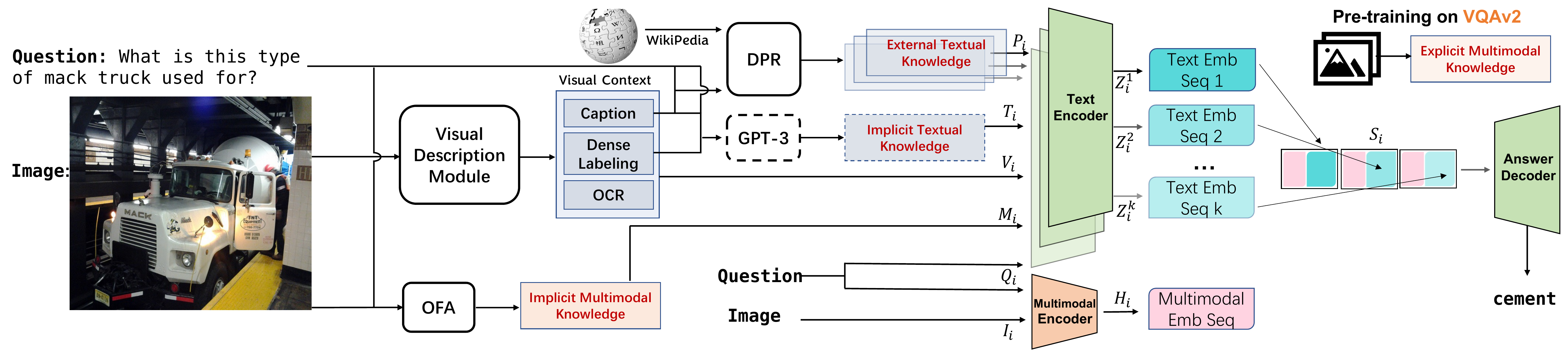} 
\caption{The flowchart of our method shows how we obtain four types of knowledge (red fonts) and feed them into the proposed model, which consists of a multimodal encoder, a textual encoder and an answer decoder. }
\label{model}
\end{figure*}

\section{Method}

\subsection{Visual Description Module}
Given an image $I_i$, following \citep{gao2022transform}, we adopt a coarse-to-fine transformation strategy to describe it as comprehensively as possible, and obtain three parts as follows. 

    1. Image-level caption $C_i$, given by the SoTA VLP model OFA \citep{wang2022ofa}.
    
    2. Object-level attribution description $L_i$ from the VinVL \citep{zhang2021vinvl} detector.
    
    3. Token-level Optical Character Recognition (OCR) results $O_i$ from easyOCR\footnote{https://github.com/JaidedAI/EasyOCR}.

To simplify, we refer to the three as visual context $V_i=(Ci, Li, Oi)$. 
The generated visual descriptions are in the following forms: 
\vspace{-0.1cm}
\begin{equation}
\begin{split} 
C_i&=\left\{ (w_0^{cap}, ... , w_j^{cap}) \right\}  \\ L_i&=\left\{ phrase_0^{lab}, ... , phrase_m^{lab} \right\}, \\
 & \ \ \ \ \ \ phrase_m^{lab} = (w_0^{attr}, ... , w_n^{attr},w^{\text {obj}}) \\O_i&=\left\{ w_0^{\text {ocr}}, ... , w_k^{\text {ocr}} \right\}   
\end{split}
\end{equation}
\vspace{-0.1cm}





\subsection{Explicit Knowledge Retrieval and Accumulation}

To answer challenging questions, humans tend to query them in knowledge bases or accumulate relevant knowledge in advance. Inspired by this, we introduce explicit textual and multimodal knowledge through retrieval and accumulation, respectively. 

\paragraph{Wikipedia Passage Retrieval.}
We view the 21-million-passage Wikipedia dump $D$ as an explicit textual knowledge source. 
In particular, 
we combine the question $Q_i$ and  caption $C_i$ as a query $q_i= (Q_i, C_i)$ to retrieve the relevant passages from $D$. 
To this end, our method adopts an off-the-shelf pre-trained dense passage retrieval (DPR) \citep{karpukhin2020dense} model. 
DPR encodes the query $q_i$ and all candidate passages $D$ separately into dense vectors $v_{q_i}$ and $[v_{p_0}, v_{p_1}, ..., v_{p_{|D|}}]$ with two independent BERT encoders as follows:
\begin{equation}
\mathbf{v}_{q_i}=BERT_Q\left(q_i\right), \mathbf{v}_{p_k}=BERT_P\left(p_k\right)
\end{equation}
We compute the inner product $sim(q_i, p_k) =\mathbf{v}_{q_i}^T \cdot \mathbf{v}_{p_k}$ as their similarity scores, and then 
exploit an indexing engine FAISS \cite{johnson2019billion} to speed up the above process. The knowledge passages $P_i=[p_{i,0}, p_{i,1}, ..., p_{i,k}]$ with top $k$ similarity scores are the final \redtxt{explicit textual knowledge}. 



\paragraph{VQA Knowledge Accumulation.}
Compared to the rigid facts of textual knowledge, the inexpressible facts of multimodal knowledge are also indispensable (e.g., object identification and scene understanding \citep{ding2022mukea}. We view the conventional VQAv2 dataset as an \redtxt{explicit multimodal knowledge} source, and our model accumulates multimodal knowledge in advance through pre-training on VQAv2. 

\subsection{Implicit Knowledge Retrieval}
Recently, the GPT-3 LLM has shown its strength in generating open domain knowledge \citep{gui2021kat,yang2022empirical} in a prompt-learning manner, and is widely used in OK-VQA 
as a source of implicit textual knowledge. 
However, the text descriptions of given images in prompts may lack important visual information, resulting in incomplete or irrelevant knowledge output from GPT-3.
To overcome such drawbacks, 
we propose to view the unifying VLP model OFA as a source of implicit multimodal knowledge. 
Different from GPT-3, OFA can be queried directly by visual features with text prompts. 


\paragraph{Implicit Textual Knowledge in GPT-3.}
Following the prompt tuning procedure of KAT \cite{gui2021kat},  
we retrieve implicit textual knowledge in GPT-3 with supporting evidence. 
Specifically, we use the combination of the question, caption, and object labeling as a prompt $X_{gpt}$ for each image-question pair. 
Then we add carefully designed instruction text and semantically similar samples as the in-context examples at the beginning of $X_{gpt}$. 
That is, $X_{gpt}$ is "$\left\langle instructions \right\rangle$ $\left\langle in-context\  
 examples\right\rangle$ \  \texttt{Context:}$\left\langle caption\ C_i \right\rangle$+$\left\langle object labeling\ L_i \right\rangle$. \ \texttt{Q:}$\left\langle question\  Q_i\right\rangle$ \ \texttt{A:}". $X_{gpt}$ can query a tentative answer $A_i^{gpt}$, and we then query GPT-3 with another prompt $Y_{gpt}$ "$\left\langle question Q_i\right\rangle \  \left\langle answer A_i^{gpt}\right\rangle$. \texttt{This is because}" for supporting evidence $E_i^{gpt}$. The final obtained \redtxt{implicit textual knowledge} is $T_i=\left\{A_i^{gpt}, E_i^{gpt}\right\}$. 


\paragraph{Implicit Multimodal Knowledge in OFA.}
Instruction-guided pre-training enables OFA to perform zero-shot generalization for different prompts, although it does not have a huge parameter capacity like GPT-3. 
To generate the tentative answer $A_i^{ofa}$, we directly feed OFA the visual features and question as the prompt $X_{gpt}$. 
In addition, "\texttt{This is because}" in $Y_{gpt}$ is no longer applicable to prompt OFA to generate the evidence, as OFA excels at question-form prompts rather than writing a continuation like GPT-3.
We therefore design
a question-form prompt $Y_{ofa}$ "$\left\langle question\ Q_i\right\rangle$ \texttt{Why} $\left\langle answer A_i^{ofa}\right\rangle$ \texttt{?}" to query OFA for supporting evidence $E_i^{ofa}$. 
The final obtained \redtxt{implicit multimodal knowledge} is $M_i=\left\{A_i^{ofa}, E_i^{ofa}\right\}$. 

\subsection{Model Structure of \ourmethod}
We have designed the modules above for different types of knowledge, and then, as shown in Figure \ref{model}, transfer the acquired knowledge to our model, which
 contains the following modules: 

\paragraph{Multimodal Encoder.}
We directly adopt an existing VLP model as our multimodal encoder. This paper mainly uses LXMERT, the most widely used one in VQA. LXMERT encodes question $Q_i$ and image $I_i$ to  
obtain the language hidden states $\hat{H_i^l}$ and vision hidden states $\hat{H_i^v}$ that have fully interacted with each other. 
\begin{equation}
    \hat{H_i^l}, \hat{H_i^v} = enc_{mm}(Q_i, I_i)
\end{equation}
where $\hat{H_i^l} \in  \mathbb{R}^{L_q*\hat{h}}$, $\hat{H_i^v} \in  \mathbb{R}^{L_v*\hat{h}}$, $L_q$ is the length of the question, $L_v$ is the number of objects, and $\hat{h}$ is the size of hidden embedding. This encoder acts like "\textit{observing}" where visual features can interact well with questions.

\paragraph{Textual Encoder.}
We use T5's encoder as the textual encoder, and feed in all possible textual information, i.e., $Q_i$, $V_i$, $M_i$(, $T_i$)\footnote{Unless compared with GPT-3 based methods,  
$T_i$ extracted from GPT-3 is not included by default, due to the much energy consumption of GPT-3.} and $P_i$ as input. 
Due to the large number of relevant Wikipedia passages, we concatenate each passage $p_{i,k}$ that iterates over $P_i$ with other inputs, and then feed each 
concatenated sequence 
into the textual encoder as: 
\begin{equation}
    Z_i^k = enc_{txt}(Q_i, V_i, M_i, p_{i,k})
\end{equation}
Here, we obtain the hidden embedding sequence $Z_i^k=(z_0, z_1, ..., z_t)$, where $z_t$ represents the $t_{th}$ token embedding, $Z_i^k \in \mathbb{R}^{L_t*h}$, $L_t = |(Q_i, V_i, M_i, p_{i,k})|$ is the length of the sequence and $h$ is the size of the hidden embedding. This encoder acts like "\textit{thinking}" where vast knowledge can interact well with questions.


\paragraph{Combo of Both Encoders.}
To combine the hidden embeddings of both encoders, we map the embedding of the multimodal encoder into the same dimensional space as the textual encoder: 
\begin{equation}
    H_i^l, H_i^v = FC_2(relu(FC_1([\hat{H_i^l}, \hat{H_i^v}])))
\end{equation}
where $H_i^l \in  \mathbb{R}^{L_q*h}$, $H_i^v \in  \mathbb{R}^{L_v*h}$. 
The final multimodal embedding sequence is $H_i = (H_i^l, H_i^v)$. 
Then we combine the multimodal and textual embedding sequence together to obtain a hybrid embedding sequence $S_i^k = (H_i, Z_i^k)$. 
Subsequently, we iterate all $k$ passages with the same encoding process to generate $k$ hybrid embedding sequences:
\begin{equation}
    S_i = (S_i^0, S_i^1, ..., S_i^k)
\end{equation} 
where $S_i \in \mathbb{R}^{((L_q+L_v+L_t)\cdot k)\times h}$ is the concatenation of all $k$ sequences. Taking into account both visual features and vast knowledge, we come to a combo of "\textit{thinking and observing}". 


\paragraph{Answer Decoder.}
We apply T5's decoder as the answer decoder, and feed in the embedding sequence $S_i$ 
to generate the final answer according to the prediction probability $P()$ over the vocabulary space $|V|$ for each answer token: 
\begin{equation}
P(a_i^1), ..., P(a_i^l)=softmax(dec(\mathbf{S_i}))
\end{equation}
where $l$ is the length of the answer. 
Finally, we adopt teacher-enforcing to train the model with auto-regressive cross-entropy objective: 
\begin{equation}
L_{ans}=\frac{-1}{N\cdot l \cdot |V|} \sum_{i=1}^N \sum_{j=1}^l \sum_{w=1}^{|V|} A_i^{j, w} \log (P(a_i^{j, w}))
\end{equation}
where $N$ is the size of the whole training set.
\paragraph{Pre-training and Fine-tuning.}
In addition to accumulating explicit multimodal knowledge in VQAv2, the pre-training stage also makes the answer decoder suitable for decoding two different encoders. 
Note that the implicit knowledge $T_i$ and $M_i$ are not used during pre-training, while the forms of other inputs are consistent with fine-tuning. 
To employ model ensemble, a common practice in OK-VQA, we take ensembles of six models trained with different seeds, and select the most frequent predictions as the final answers. 

\begin{table*}
\centering
\resizebox{0.95\linewidth}{!}{ 
\begin{tabular}{ll|c|c|ll}
\hline
\textbf{Method} & \textbf{Venue} &  \textbf{Implicit Knowledge }& \textbf{Explicit Knowledge Resources} & \textbf{EM} & \textbf{Acc}\\
\hline
BAN & \small{NeurIPS\citeyearpar{kim2018bilinear}} & ---&--- & & 25.17 \\ 
\ \ \ +AN &\small{CVPR\citeyearpar{marino2019ok}}&---&Wikipedia & & 25.61 \\
\ \ \ +KG-AUC&\small{MM\citeyearpar{li2020boosting}} &---& Wikipedia + ConceptNet & & 26.71 \\
MUTAN &\small{ICCV\citeyearpar{ben2017mutan}} & --- &--- & &26.41 \\
\ \ \ +AN &\small{CVPR\citeyearpar{marino2019ok}}&---&Wikipedia & & 27.84\\
Mucko &\small{IJCAI\citeyearpar{zhu2020mucko}}&---&ConceptNet && 29.20 \\
GRUC&\small{PR\citeyearpar{yu2020cross}}&---&ConceptNet && 29.87 \\
KM$^4$&\small{Inf Fusion\citeyearpar{zheng2021km4}}&---&multimodal knowledge from OK-VQA && 31.32 \\ \hline

ViLBERT &\small{ICNIP\citeyearpar{lu2019vilbert}} &ViLBERT&  & & 31.35 \\
LXMERT & \small{EMNLP\citeyearpar{tan2019lxmert}}&LXMERT&  & & 32.04 \\

VRR-CReader &\small{EMNLP\citeyearpar{luo2021weakly}}& LXMERT& Google Search & &36.78 \\
RVLESK &\small{LANTERN\citeyearpar{shevchenko2021reasoning}}& LXMERT& ConceptNet  & &39.04 \\
MAVEx &\small{AAAI\citeyearpar{wu2022multi}}& ViLBERT& Wikipedia + ConceptNet + Google Images & &41.37 \\
MuKEA & \small{CVPR\citeyearpar{ding2022mukea}} & LXMERT& multimodal knowledge from VQAv2 and OK-VQA & & 42.59 \\ \hline

ConceptBert& \small{EMNLP\citeyearpar{garderes2020conceptbert}} & BERT  & ConceptNet & & 33.66 \\
KRISP\small(w/o mm pre.) &\small{CVPR\citeyearpar{Marino_2021_CVPR}} &BERT& DBpedia + ConceptNet + VisualGenome + haspartKB & & 32.31 \\
KRISP\small(w/ mm pre.) &\small{CVPR\citeyearpar{Marino_2021_CVPR}} &BERT& \textbf{\textit{ditto}} + VQAv2  & & 38.90 \\
VRR-EReader &\small{EMNLP\citeyearpar{luo2021weakly}}& RoBERTa & Google Search & &39.20 \\
TRiG & \small{CVPR\citeyear{gao2022transform}}& T5& Wikipedia & 53.59 & 49.35 \\ 
TRiG, \textbf{\textit{E}}  & \small{CVPR\citeyearpar{gao2022transform}}& T5 & Wikipedia & 54.73 & 50.50 \\ \hline
Ours && LXMERT+OFA+T5 &  VQAv2 + Wikipedia &\textbf{59.85}& \textbf{55.33} \\ 
Ours, \textbf{\textit{E}}  && LXMERT+OFA+T5 & VQAv2 + Wikipedia &\textbf{61.12}&  \textbf{56.49} \\ 
Ours && visualBERT+OFA+T5 &  VQAv2 + Wikipedia &\textbf{60.17}& \textbf{55.52} \\ 
Ours, \textbf{\textit{E}}  && visualBERT+OFA+T5 & VQAv2 + Wikipedia &\textbf{61.32}&  \textbf{56.67} \\ \hline 

\hline
\end{tabular}
}
\caption{\label{main}
Results comparison with existing methods. The middle two columns report the implicit knowledge and explicit knowledge sources involved in each method respectively. The middle two rows show the methods based on VLP models and PLMs respectively. \textbf{\textit{E}} denotes the model ensemble.
}
\vspace{-0.1cm}
\end{table*}

\section{Experiments}
\subsection{Experimental Setup}
\paragraph{OK-VQA Dataset.}
This paper conducts extensive experiments on the OK-VQA dataset \cite{marino2019ok}, the most open VQA dataset, where each question requires outside knowledge beyond the image to answer correctly.
Since all questions are manually annotated with no fixed template or knowledge base, this dataset allows the use of any external knowledge source that can help answer. 

\paragraph{Evaluation Metric and Implementation Details.}
We evaluate performance by the standard VQA evaluation metric \citep{goyal2017making} (denoted by Acc) and Exact Match \citep{gao2022transform} (denoted by EM). Acc defines a soft score (between 0 and 1) for each annotated answer according to a voting mechanism, reflecting the consensus subjectively of multiple annotators. In contrast, EM treats all annotated answers to a question equally as the ground truth, which is a looser metric.

We adopt \textit{lxmert-base-uncased} or \textit{visualbert-vqa} \citep{li2019visualbert} and \textit{T5-large} models to initialize our model. 
We pre-train and finetune the models on 12 and 8 A100-80GB GPUs respectively for 3 epochs with a batch size of 1. More details are shown in Appendix \ref{moredetail}. 
 
\subsection{Comparison with Existing Approaches}

\paragraph{Comparison with SoTAs.}
Table \ref{main} reports the performance of our proposed method and state-of-the-art models, from which we can derive several observations: 
(1) Comparing the second and third lines with the first line, we find that implicit knowledge in VLP models or PLMs, used for model initialization, further improves model performance. This was rarely discussed in previous work. 
(2) MuKEA and TriG are the best-performing methods to implement OK-VQA in cross-modal space and natural-language space, respectively. 
By comparing their performance, we find that OK-VQA solutions in natural-language space perform significantly better than those in cross-modal space. 
This is because squeezing the rich representation of natural-language knowledge (billion-degree pre-training corpus) into a much smaller cross-modal space (million-degree pre-training corpus) leads to a severe loss of knowledge. 
(3) 
Our method is compatible with various VLP encoders, and beats the previous SoTAs TRiG by 6.17\% Acc and 6.59\% EM. 
(4) It can be seen from the middle two 
columns that, compared to previous work, our method is the first to utilize all four types of knowledge at the same time, which is one of the reasons why our method is effective. 
Moreover, as shown in Appendix \ref{moreWiki}, our method can outperform TRiG using 100 Wikipedia passages by 4.37\% Acc even using only 5 passages, which substantially reduces computing consumption.  

\begin{table}
\centering
\resizebox{0.9\linewidth}{!}{ 
\begin{tabular}{c|c|c}
\hline
\textbf{Method} &  \textbf{Knowledge in Input Text}& \textbf{Acc}\\
\hline
PICa & Frozen GPT-3 (175B) & 46.50 \\
PICa, \textbf{\textit{E}} &Frozen GPT-3 (175B) & 48.00 \\
KAT  &Wikidata+Frozen GPT-3 (175B)& 53.10 \\ 
KAT, \textbf{\textit{E}} &Wikidata+Frozen GPT-3 (175B) & 54.40 \\ 
REVIVE & Wikidata+Frozen GPT-3 (175B) & 56.60  \\ 
REVIVE, \textbf{\textit{E}} &  Wikidata+Frozen GPT-3 (175B) & 58.00  \\ \hline
ours &  Wikipedia+Frozen OFA (0.93B)  & 55.33 \\ 
ours, \textbf{\textit{E}}  & Wikipedia+Frozen OFA (0.93B) & 56.49\\ 
ours w/ GPT-3 &  \textbf{\textit{ditto}}+Frozen GPT-3 (175B) &\textbf{57.57} \\ 
ours w/ GPT-3, \textbf{\textit{E}}  &  \textbf{\textit{ditto}}+Frozen GPT-3 (175B)& \textbf{58.72} \\ 


\hline
\end{tabular}
}
\caption{\label{gpt-3-compare}
Results comparison with existing GPT-3 based methods. 
\textbf{\textit{E}} denotes the model ensemble. }
\end{table}

\paragraph{Comparison with GPT-3 Based Methods.}
We also compare our method with recent GPT-3 based methods. As shown in Table \ref{gpt-3-compare}, GPT-3 Based methods are significantly superior to non-GPT-3 baselines shown in Table \ref{main}. 
However, even without GPT-3 (175B), we can achieve competitive results with OFA (0.93B). 
To compare fairly, we further improve our model performance by incorporating GPT-3, and clearly surpass all GPT-3 based SoTAs. 

\subsection{Ablation Study}
\paragraph{Ablation of Pretrain-finetune Strategy. }





\begin{figure}[t]
\centering
\includegraphics[width=0.95\columnwidth]{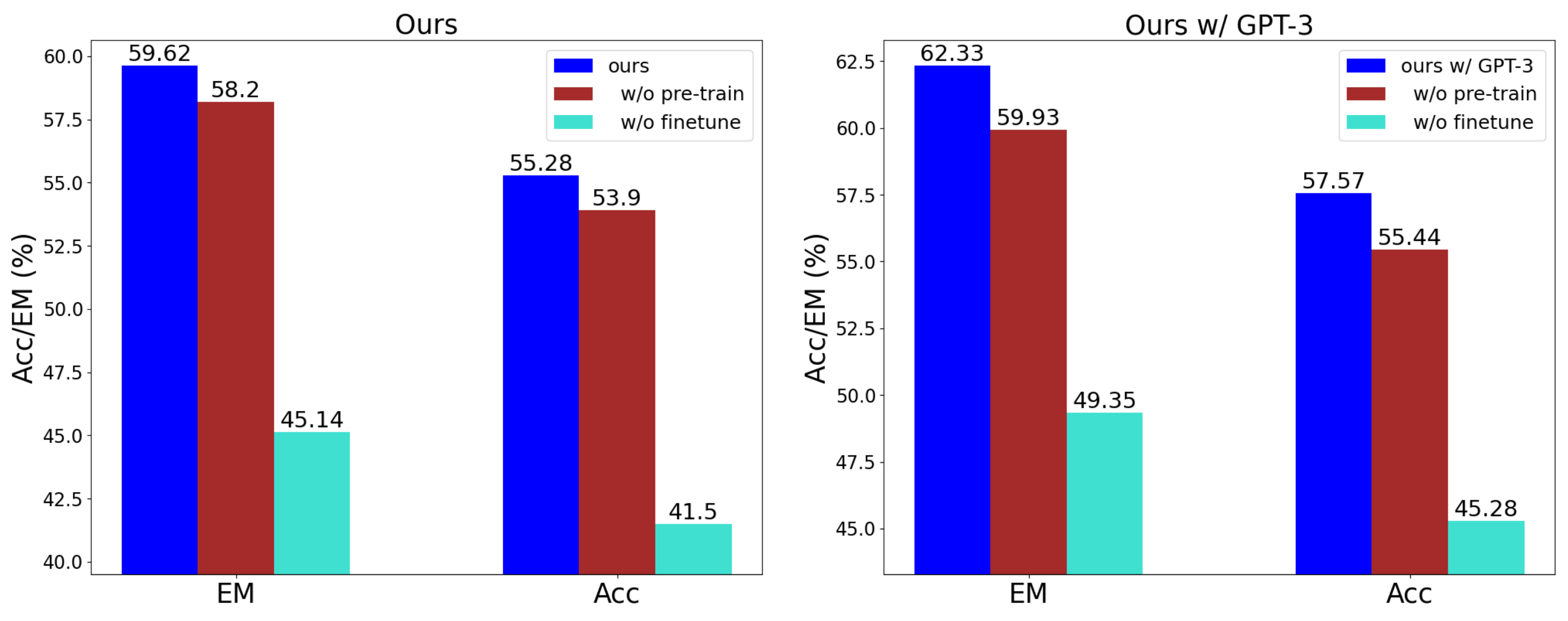} 
\vspace{-0.2cm}
\caption{Ablation study on the pre-training and fine-tuning stages. 'w/o finetune' denotes that after pre-training on VQAv2, the model will be evaluated directly on the OK-VQA test set without further fine-tuning. }
\label{pretrain}
\end{figure}

In Figure \ref{pretrain}, we evaluate the contribution of pre-training and fine-tuning in our method. 
The decline in performance caused by "w/o pre-train" confirms the necessity of pre-training. Although 'w/o finetune' is far worse than the final performance, it is still competitive compared with previous methods. This further verifies that multimodal knowledge in VQAv2 is helpful in solving OK-VQA.

\paragraph{Ablation of Model Structure.}

\begin{table}
    \centering 
    \resizebox{.95\linewidth}{!}{ 
\begin{tabular}{l|c|c|c}
\hline
\textbf{Model} & \textbf{Input Form} & \textbf{EM} & \textbf{Acc}\\
\hline
ours w/o pre. & visual features + textual input & 59.93 & 55.44 \\
\ \ \ LXMERT & visual features & --- & 32.04 \\
\ \ \ w/o txt enc & visual features & 29.43 & 26.61 \\
\ \ \ w/o mm enc & textual input & \textbf{60.01} & \textbf{55.56} \\ \hline
ours  & visual features + textual input & \textbf{62.33} & \textbf{57.57} \\
\ \ \ w/o txt enc & visual features & 34.52 & 31.39 \\
\ \ \ w/o mm enc & textual input & 61.55 & 56.83 \\

\hline
\end{tabular}
}

\caption{Ablation study on each encoder in our model structure. The middle column indicates the data format that each model can be fed. The upper part represents the models without pre-training. 'w/o txt enc' and 'w/o mm enc' denote using only multimodal encoder and textual encoder respectively. }
\label{ablationonenc}
\end{table}

To prove the complementary benefits of applying the two encoders,  
we conduct experiments and report results in Table \ref{ablationonenc}. The findings can be summarized as follows: 
(1) As shown in the "Input Form" column, combining both textual and multimodal encoders allows our method to handle both visual features and textual input simultaneously. 
(2) 'w/o txt enc' consistently underperforms  'w/o mm enc', because the natural-language space of the textual encoder contains more knowledge, which is critical to OK-VQA. 
(3) The upper part shows that, without pre-training, 'w/o textual enc' performs worse than LXMERT, as the answer decoder, initialized with T5, cannot directly fit the encoder initialized with LXMERT. 
(4) Similarly, removing the multimodal encoder without pre-training will instead result in a slight performance improvement for the same reason. 
(5) As shown in the lower part, adopting pre-training contributes to ameliorating the above phenomenon. That is, the performance of 'ours' is superior to both 'w/o txt enc' and 'w/o mm enc' by clear margins. This proves that pre-training can help make the answer decoder suitable for decoding both encoders, thus combining the advantages of both encoders. 

\paragraph{Ablation of Four Types of Knowledge. }

\begin{table}
   \centering
 \resizebox{0.9\linewidth}{!}{ 
\begin{tabular}{l|l|c|c}
\hline
\textbf{Model} & Knowledge Type& \textbf{EM} & \textbf{Acc}\\
\hline
ours &  all four types& \textbf{62.33} & \textbf{57.57} \\ \hline
\ \ \ w/o pre. &  explicit mulimodal  & 59.93 &55.44 \\
\ \ \ w/o Wiki &  explicit textual  & 60.80  &56.18\\ 
\ \ \ w/o GPT-3  & implicit mulimodal & 59.65 & 55.28 \\
\ \ \ w/o OFA & implicit  textual  & 57.13& 52.71\\
\hline
\end{tabular}
}
\caption{Ablation study on four types of knowledge. The second column lists the types of the removed knowledge source.} 
\label{knows}

\end{table}


 Table \ref{knows} shows that the absence of any type of knowledge will lead to a significant drop in performance (1.39\%\textasciitilde 4.86\% Acc and 1.53\%\textasciitilde 5.20\% EM), which proves the complementary benefits among the four types of knowledge. Among the four types of knowledge, implicit knowledge in OFA contributes the most and explicit knowledge of Wikipedia contributes the least. We will discuss this phenomenon in Appendix \ref{moreConversion}. 
 In addition, in Appendix \ref{moreAblation}, we also perform ablations from a dependence perspective to prove the indispensability of each encoder and knowledge. 












\begin{figure*}[t]
\centering
\includegraphics[width=0.99\textwidth]{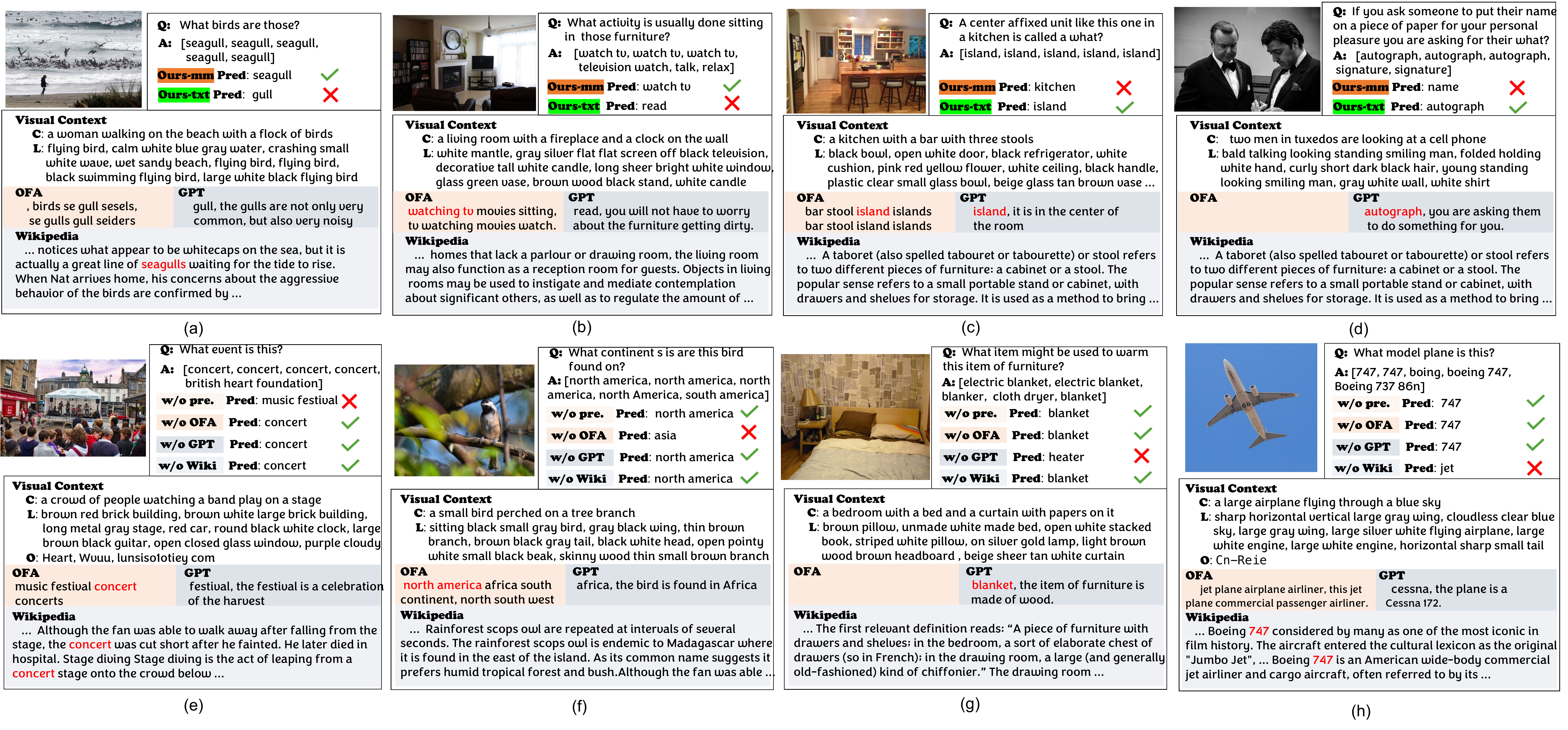} 
\vspace{-0.2cm}
\caption{ Examples of our prediction  together with all the supporting knowledge when (Upper) only using a single encoder or (Lower) respectively removing each type of knowledge from our method. \textbf{Pred} denotes our predicted answer. \textbf{ours-mm} and \textbf{ours-txt} represent the model that combines only multimodal encoder or  textual encoder with answer decoder, respectively.  
}
\vspace{-0.1cm}

\label{cases}
\end{figure*}

\begin{table}
\centering
\resizebox{.95\linewidth}{!}{ 
\begin{tabular}{l|c|c|l|c|c}
\hline

\textbf{Knowledge} & \multicolumn{2}{|c|}{\textbf{$hit$} } & \textbf{Knowledge} & \multicolumn{2}{|c}{\textbf{$hit$}} \\
\textbf{ Source}& Train & Test &\textbf{Source} & Train &Test \\ \hline

\textbf{GPT-3 ans + evi} &\textbf{56.59} &\textbf{61.51}& \textbf{OFA ans + evi} &\textbf{63.36}&\textbf{66.75}\\
GPT-3 ans  &54.02 &59.27& OFA ans  &57.63&61.59\\
GPT-3 evi &34.09 & 37.26& OFA evi &57.84&61.47\\ \hline
\textbf{Visual Context} &\textbf{32.28} & \textbf{32.9}2& \textbf{Wikipedia(75)}&\textbf{82.58} &\textbf{85.26}\\
captions &22.34 & 22.81& Wikipedia(50)&80.34 &82.62\\
labels & 23.62& 24.18& Wikipedia(25)&74.28 &76.56\\
OCR & 0.44& 0.32& Wikipedia(10)&63.20 &64.74\\ \cline{1-3}
\textbf{all} & \textbf{93.18} & \textbf{95.30} & Wikipedia(5)&51.88&54.12\\
\hline
\end{tabular}
}
\caption{\label{hit}
$Hit$ of each component in our model's inputs. $Hit$ is defined as the percentage of samples in the whole dataset that get a $hit$ on any corresponding annotated answer by the retrieved knowledge. "ans" and "evi" denote tentative answers and supporting evidence, respectively.
}
\end{table}

\paragraph{Performance of Knowledge Retrieval.}

From Table \ref{hit}, it can be seen that: 
(1) The combination of all the  knowledge retrieved in our method can cover the answers corresponding to 95.30\% of the samples.  The high $hit$ guarantees a high upper bound, allowing the model to generalize better. 
(2) $Hit$ of prompting OFA  significantly outperforms that of prompting GPT-3, indicating that implicit multimodal knowledge may be more effective than implicit textual knowledge in OK-VQA. 
(3) The supporting evidence can clearly improve $hit$ of the tentative answers, especially for OFA (from 61.59\% to 66.75\%). 
(4) Wikipedia's  high $hit$  demonstrates the effectiveness of our adopted DPR model in retrieval. As shown in  Appendix \ref{moreWiki}, 
as the number of Wikipedia passages increases, Acc/EM of our model rises first and then falls because noise is introduced when the number of passages is large. 

In Appendix \ref{moreConversion}, we also conduct experiments to further explore the extent to which the model makes use of each type of knowledge. We find that compared with explicit knowledge, implicit knowledge has a higher conversion rate from knowledge to correct answers. We also qualitatively analyze the impact on OK-VQA of different versions of OFA in Appendix \ref{moreOFA}.

\section{ Qualitative  Analysis}

\paragraph{Case Study on Two Encoders.}
To explore the respective roles of the two encoders, the upper part of Figure \ref{cases} shows the examples that can be answered correctly by one of the two single-encoder models.
 Plot (a) and (b) of Figure \ref{cases} show that \textbf{ours-mm}  excels at answering questions that need comprehension about image scenes and objects. For example, the orientation and the relative position between TV and sofa in plot (b) help generate the answer "watch tv". 
 Such scene information is easily omitted by a single textual encoder. This further validates that the multimodal encoder supplements the missing image information, and makes better use of the image when combining knowledge. 
 
 
 Plot (c) and (d) shows that \textbf{ours-txt} is an expert in answering questions that require focusing more on external knowledge rather than image understanding, since the textual encoder is the primary channel for receiving knowledge from multiple sources. 
 
\paragraph{Case Study on Varying Types of Knowledge.}
As shown in the lower plots in Figure \ref{cases}, we further analyze the circumstances under which each type of knowledge is essential, respectively. Plot (e) shows that the model would hardly generate correct answers, even those that have been recalled by knowledge, once pre-training is removed. This demonstrates that explicit multimodal knowledge accumulated during pre-training enhances the ability to use the recalled knowledge according to image content. 
Plot (f) shows that when a question is deeply dependent on image content (e.g., bird type detection), implicit multimodal knowledge in OFA can directly provide tentative answers from the image, which strengthens the visual understanding. 
Plot (g) shows that implicit textual knowledge in GPT-3 is essential for questions that require commonsense knowledge. Plot (h) shows that when a question is highly open, even if both GPT-3 and OFA fail to recall the corresponding knowledge, the retrieved Wikipedia passage can still provide enough knowledge (see Figure \ref{knows}), e.g., enumerating the most plane models. In Appendix \ref{morecase}, we also compare our method qualitatively against the previous methods.


\section{Conclusion and Future Work}
This paper proposes a simple and effective method that mimics human behavior "\textit{thinking while observing}", i.e.,  benefiting from the vast knowledge in natural-language space while making the most of the visual features for better image understanding. Our method establishes a new SoTA accuracy of 56.67\% with a 6.17\% improvement on OK-VQA. 
Moreover, we consider more comprehensive types of knowledge, and systematically analyze the role of each type of knowledge in detail. We hope our work can stimulate followers to explore OK-VQA further along the direction of how to fuse both nature-language and cross-modality spaces better.

\section*{Limitations}
First, considering the unique most open setting of OK-VQA, following  
most previous work \citep{gao2022transform,wu2022multi,yang2022empirical,gui2021kat,lin2022revive}, we only evaluate our method on this dataset. 
Second, although the proposed method has verified the feasibility of the idea that constrains both natural-language and cross-modality spaces together, it is still necessary to explore more ways to better combine the output of two encoders. 
Third, our method involves multiple offline knowledge retrieval processes, such as retrieving relevant Wikipedia passages, which will make it difficult to deploy our model as an online model. 

%
\bibliography{anthology,custom}

\begin{thebibliography}{62}
\expandafter\ifx\csname natexlab\endcsname\relax\def\natexlab#1{#1}\fi

\bibitem[{Agrawal et~al.(2016)Agrawal, Batra, and
  Parikh}]{agrawal2016analyzing}
Aishwarya Agrawal, Dhruv Batra, and Devi Parikh. 2016.
\newblock Analyzing the behavior of visual question answering models.
\newblock \emph{arXiv preprint arXiv:1606.07356}.

\bibitem[{Agrawal et~al.(2018)Agrawal, Batra, Parikh, and
  Kembhavi}]{agrawal2018don}
Aishwarya Agrawal, Dhruv Batra, Devi Parikh, and Aniruddha Kembhavi. 2018.
\newblock Don't just assume; look and answer: Overcoming priors for visual
  question answering.
\newblock In \emph{Proceedings of the IEEE conference on computer vision and
  pattern recognition}, pages 4971--4980.

\bibitem[{Alayrac et~al.(2022)Alayrac, Donahue, Luc, Miech, Barr, Hasson, Lenc,
  Mensch, Millican, Reynolds et~al.}]{alayrac2022flamingo}
Jean-Baptiste Alayrac, Jeff Donahue, Pauline Luc, Antoine Miech, Iain Barr,
  Yana Hasson, Karel Lenc, Arthur Mensch, Katie Millican, Malcolm Reynolds,
  et~al. 2022.
\newblock Flamingo: a visual language model for few-shot learning.
\newblock \emph{arXiv preprint arXiv:2204.14198}.

\bibitem[{Antol et~al.(2015)Antol, Agrawal, Lu, Mitchell, Batra, Zitnick, and
  Parikh}]{antol2015vqa}
Stanislaw Antol, Aishwarya Agrawal, Jiasen Lu, Margaret Mitchell, Dhruv Batra,
  C~Lawrence Zitnick, and Devi Parikh. 2015.
\newblock Vqa: Visual question answering.
\newblock In \emph{Proceedings of the IEEE international conference on computer
  vision}, pages 2425--2433.

\bibitem[{Baudi{\v{s}} and {\v{S}}ediv{\`y}(2015)}]{baudivs2015modeling}
Petr Baudi{\v{s}} and Jan {\v{S}}ediv{\`y}. 2015.
\newblock Modeling of the question answering task in the yodaqa system.
\newblock In \emph{International Conference of the cross-language evaluation
  Forum for European languages}, pages 222--228. Springer.

\bibitem[{Ben-Younes et~al.(2017)Ben-Younes, Cadene, Cord, and
  Thome}]{ben2017mutan}
Hedi Ben-Younes, R{\'e}mi Cadene, Matthieu Cord, and Nicolas Thome. 2017.
\newblock Mutan: Multimodal tucker fusion for visual question answering.
\newblock In \emph{Proceedings of the IEEE international conference on computer
  vision}, pages 2612--2620.

\bibitem[{Berant et~al.(2013)Berant, Chou, Frostig, and
  Liang}]{berant2013semantic}
Jonathan Berant, Andrew Chou, Roy Frostig, and Percy Liang. 2013.
\newblock Semantic parsing on freebase from question-answer pairs.
\newblock In \emph{Proceedings of the 2013 conference on empirical methods in
  natural language processing}, pages 1533--1544.

\bibitem[{Brown et~al.(2020)Brown, Mann, Ryder, Subbiah, Kaplan, Dhariwal,
  Neelakantan, Shyam, Sastry, Askell et~al.}]{brown2020language}
Tom Brown, Benjamin Mann, Nick Ryder, Melanie Subbiah, Jared~D Kaplan, Prafulla
  Dhariwal, Arvind Neelakantan, Pranav Shyam, Girish Sastry, Amanda Askell,
  et~al. 2020.
\newblock Language models are few-shot learners.
\newblock \emph{Advances in neural information processing systems},
  33:1877--1901.

\bibitem[{Cao et~al.(2021)Cao, Li, Liang, Wang, and Lin}]{cao2021knowledge}
Qingxing Cao, Bailin Li, Xiaodan Liang, Keze Wang, and Liang Lin. 2021.
\newblock Knowledge-routed visual question reasoning: Challenges for deep
  representation embedding.
\newblock \emph{IEEE Transactions on Neural Networks and Learning Systems}.

\bibitem[{Chen et~al.(2017)Chen, Fisch, Weston, and Bordes}]{chen2017reading}
Danqi Chen, Adam Fisch, Jason Weston, and Antoine Bordes. 2017.
\newblock Reading wikipedia to answer open-domain questions.
\newblock \emph{arXiv preprint arXiv:1704.00051}.

\bibitem[{Chen et~al.(2022)Chen, Wang, Changpinyo, Piergiovanni, Padlewski,
  Salz, Goodman, Grycner, Mustafa, Beyer et~al.}]{chen2022pali}
Xi~Chen, Xiao Wang, Soravit Changpinyo, AJ~Piergiovanni, Piotr Padlewski,
  Daniel Salz, Sebastian Goodman, Adam Grycner, Basil Mustafa, Lucas Beyer,
  et~al. 2022.
\newblock Pali: A jointly-scaled multilingual language-image model.
\newblock \emph{arXiv preprint arXiv:2209.06794}.

\bibitem[{Chen et~al.(2019)Chen, Li, Yu, El~Kholy, Ahmed, Gan, Cheng, and
  Liu}]{chen2019uniter}
Yen-Chun Chen, Linjie Li, Licheng Yu, Ahmed El~Kholy, Faisal Ahmed, Zhe Gan,
  Yu~Cheng, and Jingjing Liu. 2019.
\newblock Uniter: Learning universal image-text representations.

\bibitem[{Dancette et~al.(2021)Dancette, Cadene, Teney, and
  Cord}]{dancette2021beyond}
Corentin Dancette, Remi Cadene, Damien Teney, and Matthieu Cord. 2021.
\newblock Beyond question-based biases: Assessing multimodal shortcut learning
  in visual question answering.
\newblock In \emph{Proceedings of the IEEE/CVF International Conference on
  Computer Vision}, pages 1574--1583.

\bibitem[{Devlin et~al.(2018)Devlin, Chang, Lee, and
  Toutanova}]{devlin2018bert}
Jacob Devlin, Ming-Wei Chang, Kenton Lee, and Kristina Toutanova. 2018.
\newblock Bert: Pre-training of deep bidirectional transformers for language
  understanding.
\newblock \emph{arXiv preprint arXiv:1810.04805}.

\bibitem[{Ding et~al.(2022)Ding, Yu, Liu, Hu, Cui, and Wu}]{ding2022mukea}
Yang Ding, Jing Yu, Bang Liu, Yue Hu, Mingxin Cui, and Qi~Wu. 2022.
\newblock Mukea: Multimodal knowledge extraction and accumulation for
  knowledge-based visual question answering.
\newblock In \emph{Proceedings of the IEEE/CVF Conference on Computer Vision
  and Pattern Recognition}, pages 5089--5098.

\bibitem[{Gao et~al.(2022)Gao, Ping, Thattai, Reganti, Wu, and
  Natarajan}]{gao2022transform}
Feng Gao, Qing Ping, Govind Thattai, Aishwarya Reganti, Ying~Nian Wu, and Prem
  Natarajan. 2022.
\newblock Transform-retrieve-generate: Natural language-centric
  outside-knowledge visual question answering.
\newblock In \emph{Proceedings of the IEEE/CVF Conference on Computer Vision
  and Pattern Recognition}, pages 5067--5077.

\bibitem[{Gard{\`e}res et~al.(2020)Gard{\`e}res, Ziaeefard, Abeloos, and
  Lecue}]{garderes2020conceptbert}
Fran{\c{c}}ois Gard{\`e}res, Maryam Ziaeefard, Baptiste Abeloos, and Freddy
  Lecue. 2020.
\newblock Conceptbert: Concept-aware representation for visual question
  answering.
\newblock In \emph{Findings of the Association for Computational Linguistics:
  EMNLP 2020}, pages 489--498.

\bibitem[{Goyal et~al.(2017)Goyal, Khot, Summers-Stay, Batra, and
  Parikh}]{goyal2017making}
Yash Goyal, Tejas Khot, Douglas Summers-Stay, Dhruv Batra, and Devi Parikh.
  2017.
\newblock Making the v in vqa matter: Elevating the role of image understanding
  in visual question answering.
\newblock In \emph{Proceedings of the IEEE conference on computer vision and
  pattern recognition}, pages 6904--6913.

\bibitem[{Gui et~al.(2021)Gui, Wang, Huang, Hauptmann, Bisk, and
  Gao}]{gui2021kat}
Liangke Gui, Borui Wang, Qiuyuan Huang, Alex Hauptmann, Yonatan Bisk, and
  Jianfeng Gao. 2021.
\newblock Kat: A knowledge augmented transformer for vision-and-language.
\newblock \emph{arXiv preprint arXiv:2112.08614}.

\bibitem[{Guo et~al.(2021)Guo, Xu, and Tao}]{guo2021bilinear}
Dalu Guo, Chang Xu, and Dacheng Tao. 2021.
\newblock Bilinear graph networks for visual question answering.
\newblock \emph{IEEE Transactions on Neural Networks and Learning Systems}.

\bibitem[{Hirota et~al.(2022)Hirota, Nakashima, and Garcia}]{hirota2022gender}
Yusuke Hirota, Yuta Nakashima, and Noa Garcia. 2022.
\newblock Gender and racial bias in visual question answering datasets.
\newblock \emph{arXiv preprint arXiv:2205.08148}.

\bibitem[{Hudson and Manning(2019)}]{hudson2019gqa}
Drew~A Hudson and Christopher~D Manning. 2019.
\newblock Gqa: A new dataset for real-world visual reasoning and compositional
  question answering.
\newblock In \emph{Proceedings of the IEEE/CVF conference on computer vision
  and pattern recognition}, pages 6700--6709.

\bibitem[{Jiang et~al.(2020)Jiang, Misra, Rohrbach, Learned-Miller, and
  Chen}]{jiang2020defense}
Huaizu Jiang, Ishan Misra, Marcus Rohrbach, Erik Learned-Miller, and Xinlei
  Chen. 2020.
\newblock In defense of grid features for visual question answering.
\newblock In \emph{Proceedings of the IEEE/CVF Conference on Computer Vision
  and Pattern Recognition}, pages 10267--10276.

\bibitem[{Johnson et~al.(2019)Johnson, Douze, and
  J{\'e}gou}]{johnson2019billion}
Jeff Johnson, Matthijs Douze, and Herv{\'e} J{\'e}gou. 2019.
\newblock Billion-scale similarity search with gpus.
\newblock \emph{IEEE Transactions on Big Data}, 7(3):535--547.

\bibitem[{Johnson et~al.(2017)Johnson, Hariharan, Van Der~Maaten, Fei-Fei,
  Lawrence~Zitnick, and Girshick}]{johnson2017clevr}
Justin Johnson, Bharath Hariharan, Laurens Van Der~Maaten, Li~Fei-Fei,
  C~Lawrence~Zitnick, and Ross Girshick. 2017.
\newblock Clevr: A diagnostic dataset for compositional language and elementary
  visual reasoning.
\newblock In \emph{Proceedings of the IEEE conference on computer vision and
  pattern recognition}, pages 2901--2910.

\bibitem[{Joshi et~al.(2017)Joshi, Choi, Weld, and
  Zettlemoyer}]{joshi2017triviaqa}
Mandar Joshi, Eunsol Choi, Daniel~S Weld, and Luke Zettlemoyer. 2017.
\newblock Triviaqa: A large scale distantly supervised challenge dataset for
  reading comprehension.
\newblock \emph{arXiv preprint arXiv:1705.03551}.

\bibitem[{Karpukhin et~al.(2020)Karpukhin, O{\u{g}}uz, Min, Lewis, Wu, Edunov,
  Chen, and Yih}]{karpukhin2020dense}
Vladimir Karpukhin, Barlas O{\u{g}}uz, Sewon Min, Patrick Lewis, Ledell Wu,
  Sergey Edunov, Danqi Chen, and Wen-tau Yih. 2020.
\newblock Dense passage retrieval for open-domain question answering.
\newblock \emph{arXiv preprint arXiv:2004.04906}.

\bibitem[{Kim et~al.(2018)Kim, Jun, and Zhang}]{kim2018bilinear}
Jin-Hwa Kim, Jaehyun Jun, and Byoung-Tak Zhang. 2018.
\newblock Bilinear attention networks.
\newblock \emph{Advances in neural information processing systems}, 31.

\bibitem[{Krishna et~al.(2017)Krishna, Zhu, Groth, Johnson, Hata, Kravitz,
  Chen, Kalantidis, Li, Shamma et~al.}]{krishna2017visual}
Ranjay Krishna, Yuke Zhu, Oliver Groth, Justin Johnson, Kenji Hata, Joshua
  Kravitz, Stephanie Chen, Yannis Kalantidis, Li-Jia Li, David~A Shamma, et~al.
  2017.
\newblock Visual genome: Connecting language and vision using crowdsourced
  dense image annotations.
\newblock \emph{International journal of computer vision}, 123(1):32--73.

\bibitem[{Kwiatkowski et~al.(2019)Kwiatkowski, Palomaki, Redfield, Collins,
  Parikh, Alberti, Epstein, Polosukhin, Devlin, Lee
  et~al.}]{kwiatkowski2019natural}
Tom Kwiatkowski, Jennimaria Palomaki, Olivia Redfield, Michael Collins, Ankur
  Parikh, Chris Alberti, Danielle Epstein, Illia Polosukhin, Jacob Devlin,
  Kenton Lee, et~al. 2019.
\newblock Natural questions: a benchmark for question answering research.
\newblock \emph{Transactions of the Association for Computational Linguistics},
  7:453--466.

\bibitem[{Lee et~al.(2019)Lee, Chang, and Toutanova}]{lee2019latent}
Kenton Lee, Ming-Wei Chang, and Kristina Toutanova. 2019.
\newblock Latent retrieval for weakly supervised open domain question
  answering.
\newblock \emph{arXiv preprint arXiv:1906.00300}.

\bibitem[{Li et~al.(2020{\natexlab{a}})Li, Wang, and Zhu}]{li2020boosting}
Guohao Li, Xin Wang, and Wenwu Zhu. 2020{\natexlab{a}}.
\newblock Boosting visual question answering with context-aware knowledge
  aggregation.
\newblock In \emph{Proceedings of the 28th ACM International Conference on
  Multimedia}, pages 1227--1235.

\bibitem[{Li et~al.(2019)Li, Yatskar, Yin, Hsieh, and Chang}]{li2019visualbert}
Liunian~Harold Li, Mark Yatskar, Da~Yin, Cho-Jui Hsieh, and Kai-Wei Chang.
  2019.
\newblock Visualbert: A simple and performant baseline for vision and language.
\newblock \emph{arXiv preprint arXiv:1908.03557}.

\bibitem[{Li et~al.(2020{\natexlab{b}})Li, Gao, Niu, Xiao, Liu, Liu, Wu, and
  Wang}]{li2020unimo}
Wei Li, Can Gao, Guocheng Niu, Xinyan Xiao, Hao Liu, Jiachen Liu, Hua Wu, and
  Haifeng Wang. 2020{\natexlab{b}}.
\newblock Unimo: Towards unified-modal understanding and generation via
  cross-modal contrastive learning.
\newblock \emph{arXiv preprint arXiv:2012.15409}.

\bibitem[{Lin et~al.(2022)Lin, Xie, Chen, Xu, Zhu, and Yuan}]{lin2022revive}
Yuanze Lin, Yujia Xie, Dongdong Chen, Yichong Xu, Chenguang Zhu, and Lu~Yuan.
  2022.
\newblock Revive: Regional visual representation matters in knowledge-based
  visual question answering.
\newblock \emph{arXiv preprint arXiv:2206.01201}.

\bibitem[{Liu et~al.(2019)Liu, Ott, Goyal, Du, Joshi, Chen, Levy, Lewis,
  Zettlemoyer, and Stoyanov}]{liu2019roberta}
Yinhan Liu, Myle Ott, Naman Goyal, Jingfei Du, Mandar Joshi, Danqi Chen, Omer
  Levy, Mike Lewis, Luke Zettlemoyer, and Veselin Stoyanov. 2019.
\newblock Roberta: A robustly optimized bert pretraining approach.
\newblock \emph{arXiv preprint arXiv:1907.11692}.

\bibitem[{Loshchilov and Hutter(2017)}]{loshchilov2017decoupled}
Ilya Loshchilov and Frank Hutter. 2017.
\newblock Decoupled weight decay regularization.
\newblock \emph{arXiv preprint arXiv:1711.05101}.

\bibitem[{Lu et~al.(2019)Lu, Batra, Parikh, and Lee}]{lu2019vilbert}
Jiasen Lu, Dhruv Batra, Devi Parikh, and Stefan Lee. 2019.
\newblock Vilbert: Pretraining task-agnostic visiolinguistic representations
  for vision-and-language tasks.
\newblock \emph{Advances in neural information processing systems}, 32.

\bibitem[{Lu et~al.(2018)Lu, Ji, Zhang, Duan, Zhou, and Wang}]{lu2018r}
Pan Lu, Lei Ji, Wei Zhang, Nan Duan, Ming Zhou, and Jianyong Wang. 2018.
\newblock R-vqa: learning visual relation facts with semantic attention for
  visual question answering.
\newblock In \emph{Proceedings of the 24th ACM SIGKDD International Conference
  on Knowledge Discovery \& Data Mining}, pages 1880--1889.

\bibitem[{Luo et~al.(2021)Luo, Zeng, Banerjee, and Baral}]{luo2021weakly}
Man Luo, Yankai Zeng, Pratyay Banerjee, and Chitta Baral. 2021.
\newblock Weakly-supervised visual-retriever-reader for knowledge-based
  question answering.
\newblock \emph{arXiv preprint arXiv:2109.04014}.

\bibitem[{Manjunatha et~al.(2019)Manjunatha, Saini, and
  Davis}]{manjunatha2019explicit}
Varun Manjunatha, Nirat Saini, and Larry~S Davis. 2019.
\newblock Explicit bias discovery in visual question answering models.
\newblock In \emph{Proceedings of the IEEE/CVF Conference on Computer Vision
  and Pattern Recognition}, pages 9562--9571.

\bibitem[{Marino et~al.(2021)Marino, Chen, Parikh, Gupta, and
  Rohrbach}]{Marino_2021_CVPR}
Kenneth Marino, Xinlei Chen, Devi Parikh, Abhinav Gupta, and Marcus Rohrbach.
  2021.
\newblock Krisp: Integrating implicit and symbolic knowledge for open-domain
  knowledge-based vqa.
\newblock In \emph{Proceedings of the IEEE/CVF Conference on Computer Vision
  and Pattern Recognition (CVPR)}, pages 14111--14121.

\bibitem[{Marino et~al.(2019)Marino, Rastegari, Farhadi, and
  Mottaghi}]{marino2019ok}
Kenneth Marino, Mohammad Rastegari, Ali Farhadi, and Roozbeh Mottaghi. 2019.
\newblock Ok-vqa: A visual question answering benchmark requiring external
  knowledge.
\newblock In \emph{Proceedings of the IEEE/cvf conference on computer vision
  and pattern recognition}, pages 3195--3204.

\bibitem[{Raffel et~al.(2020)Raffel, Shazeer, Roberts, Lee, Narang, Matena,
  Zhou, Li, Liu et~al.}]{raffel2020exploring}
Colin Raffel, Noam Shazeer, Adam Roberts, Katherine Lee, Sharan Narang, Michael
  Matena, Yanqi Zhou, Wei Li, Peter~J Liu, et~al. 2020.
\newblock Exploring the limits of transfer learning with a unified text-to-text
  transformer.
\newblock \emph{J. Mach. Learn. Res.}, 21(140):1--67.

\bibitem[{Shah et~al.(2019)Shah, Mishra, Yadati, and Talukdar}]{shah2019kvqa}
Sanket Shah, Anand Mishra, Naganand Yadati, and Partha~Pratim Talukdar. 2019.
\newblock Kvqa: Knowledge-aware visual question answering.
\newblock In \emph{Proceedings of the AAAI conference on artificial
  intelligence}, volume~33, pages 8876--8884.

\bibitem[{Shevchenko et~al.(2021)Shevchenko, Teney, Dick, and
  Hengel}]{shevchenko2021reasoning}
Violetta Shevchenko, Damien Teney, Anthony Dick, and Anton van~den Hengel.
  2021.
\newblock Reasoning over vision and language: Exploring the benefits of
  supplemental knowledge.
\newblock \emph{arXiv preprint arXiv:2101.06013}.

\bibitem[{Si et~al.(2022)Si, Meng, Zheng, Lin, Liu, Fu, Cao, Wang, and
  Zhou}]{si2022language}
Qingyi Si, Fandong Meng, Mingyu Zheng, Zheng Lin, Yuanxin Liu, Peng Fu, Yanan
  Cao, Weiping Wang, and Jie Zhou. 2022.
\newblock Language prior is not the only shortcut: A benchmark for shortcut
  learning in vqa.
\newblock \emph{arXiv preprint arXiv:2210.04692}.

\bibitem[{Singh et~al.(2019)Singh, Natarajan, Shah, Jiang, Chen, Batra, Parikh,
  and Rohrbach}]{singh2019towards}
Amanpreet Singh, Vivek Natarajan, Meet Shah, Yu~Jiang, Xinlei Chen, Dhruv
  Batra, Devi Parikh, and Marcus Rohrbach. 2019.
\newblock Towards vqa models that can read.
\newblock In \emph{Proceedings of the IEEE/CVF conference on computer vision
  and pattern recognition}, pages 8317--8326.

\bibitem[{Speer et~al.(2017)Speer, Chin, and Havasi}]{speer2017conceptnet}
Robyn Speer, Joshua Chin, and Catherine Havasi. 2017.
\newblock Conceptnet 5.5: An open multilingual graph of general knowledge.
\newblock In \emph{Thirty-first AAAI conference on artificial intelligence}.

\bibitem[{Tan and Bansal(2019)}]{tan2019lxmert}
Hao Tan and Mohit Bansal. 2019.
\newblock Lxmert: Learning cross-modality encoder representations from
  transformers.
\newblock \emph{arXiv preprint arXiv:1908.07490}.

\bibitem[{Tapaswi et~al.(2016)Tapaswi, Zhu, Stiefelhagen, Torralba, Urtasun,
  and Fidler}]{tapaswi2016movieqa}
Makarand Tapaswi, Yukun Zhu, Rainer Stiefelhagen, Antonio Torralba, Raquel
  Urtasun, and Sanja Fidler. 2016.
\newblock Movieqa: Understanding stories in movies through question-answering.
\newblock In \emph{Proceedings of the IEEE conference on computer vision and
  pattern recognition}, pages 4631--4640.

\bibitem[{Wang et~al.(2017)Wang, Wu, Shen, Dick, and Van
  Den~Hengel}]{wang2017fvqa}
Peng Wang, Qi~Wu, Chunhua Shen, Anthony Dick, and Anton Van Den~Hengel. 2017.
\newblock Fvqa: Fact-based visual question answering.
\newblock \emph{IEEE transactions on pattern analysis and machine
  intelligence}, 40(10):2413--2427.

\bibitem[{Wang et~al.(2015)Wang, Wu, Shen, Hengel, and Dick}]{wang2015explicit}
Peng Wang, Qi~Wu, Chunhua Shen, Anton van~den Hengel, and Anthony Dick. 2015.
\newblock Explicit knowledge-based reasoning for visual question answering.
\newblock \emph{arXiv preprint arXiv:1511.02570}.

\bibitem[{Wang et~al.(2022)Wang, Yang, Men, Lin, Bai, Li, Ma, Zhou, Zhou, and
  Yang}]{wang2022ofa}
Peng Wang, An~Yang, Rui Men, Junyang Lin, Shuai Bai, Zhikang Li, Jianxin Ma,
  Chang Zhou, Jingren Zhou, and Hongxia Yang. 2022.
\newblock Ofa: Unifying architectures, tasks, and modalities through a simple
  sequence-to-sequence learning framework.
\newblock In \emph{International Conference on Machine Learning}, pages
  23318--23340. PMLR.

\bibitem[{Wolf et~al.(2020)Wolf, Debut, Sanh, Chaumond, Delangue, Moi, Cistac,
  Rault, Louf, Funtowicz et~al.}]{wolf2020transformers}
Thomas Wolf, Lysandre Debut, Victor Sanh, Julien Chaumond, Clement Delangue,
  Anthony Moi, Pierric Cistac, Tim Rault, R{\'e}mi Louf, Morgan Funtowicz,
  et~al. 2020.
\newblock Transformers: State-of-the-art natural language processing.
\newblock In \emph{Proceedings of the 2020 conference on empirical methods in
  natural language processing: system demonstrations}, pages 38--45.

\bibitem[{Wu et~al.(2022)Wu, Lu, Sabharwal, and Mottaghi}]{wu2022multi}
Jialin Wu, Jiasen Lu, Ashish Sabharwal, and Roozbeh Mottaghi. 2022.
\newblock Multi-modal answer validation for knowledge-based vqa.
\newblock In \emph{Proceedings of the AAAI Conference on Artificial
  Intelligence}, volume~36, pages 2712--2721.

\bibitem[{Yang et~al.(2022)Yang, Gan, Wang, Hu, Lu, Liu, and
  Wang}]{yang2022empirical}
Zhengyuan Yang, Zhe Gan, Jianfeng Wang, Xiaowei Hu, Yumao Lu, Zicheng Liu, and
  Lijuan Wang. 2022.
\newblock An empirical study of gpt-3 for few-shot knowledge-based vqa.
\newblock In \emph{Proceedings of the AAAI Conference on Artificial
  Intelligence}, volume~36, pages 3081--3089.

\bibitem[{Yu et~al.(2020)Yu, Zhu, Wang, Zhang, Hu, and Tan}]{yu2020cross}
Jing Yu, Zihao Zhu, Yujing Wang, Weifeng Zhang, Yue Hu, and Jianlong Tan. 2020.
\newblock Cross-modal knowledge reasoning for knowledge-based visual question
  answering.
\newblock \emph{Pattern Recognition}, 108:107563.

\bibitem[{Yu et~al.(2019)Yu, Yu, Cui, Tao, and Tian}]{yu2019deep}
Zhou Yu, Jun Yu, Yuhao Cui, Dacheng Tao, and Qi~Tian. 2019.
\newblock Deep modular co-attention networks for visual question answering.
\newblock In \emph{Proceedings of the IEEE/CVF conference on computer vision
  and pattern recognition}, pages 6281--6290.

\bibitem[{Zhang et~al.(2021)Zhang, Li, Hu, Yang, Zhang, Wang, Choi, and
  Gao}]{zhang2021vinvl}
Pengchuan Zhang, Xiujun Li, Xiaowei Hu, Jianwei Yang, Lei Zhang, Lijuan Wang,
  Yejin Choi, and Jianfeng Gao. 2021.
\newblock Vinvl: Making visual representations matter in vision-language
  models.

\bibitem[{Zheng et~al.(2021)Zheng, Yan, Gou, and Wang}]{zheng2021km4}
Wenbo Zheng, Lan Yan, Chao Gou, and Fei-Yue Wang. 2021.
\newblock Km4: visual reasoning via knowledge embedding memory model with
  mutual modulation.
\newblock \emph{Information Fusion}, 67:14--28.

\bibitem[{Zhu et~al.(2020)Zhu, Yu, Wang, Sun, Hu, and Wu}]{zhu2020mucko}
Zihao Zhu, Jing Yu, Yujing Wang, Yajing Sun, Yue Hu, and Qi~Wu. 2020.
\newblock Mucko: multi-layer cross-modal knowledge reasoning for fact-based
  visual question answering.
\newblock \emph{arXiv preprint arXiv:2006.09073}.

\end{thebibliography}
\bibliographystyle{acl_natbib}

\appendix

\section{More Related Work}
\subsection{Relationship with Previous Works}\label{morerelation}

TRiG \citep{gao2022transform} and MuKEA \citep{ding2022mukea} respectively explored how to solve OK-VQA in natural language space and cross-modality space. 
The difference between our work and these two work can be explained by Figure \ref{paradigm}. 
KAT \citep{gui2021kat} studied two types of knowledge, i.e., implicit and explicit knowledge in natural-language space. 
We further introduced four specific types of knowledge, i.e., implicit textual and multimodal knowledge, and explicit textual and multimodal knowledge. 

Although REVIVE \citep{lin2022revive} integrates visual features into the final model as we did, their model structure and knowledge introduction strategy are different from ours. For the model structure, they connect CLIP and T5 in series (i.e., feeding T5 with visual features obtained by CLIP) while we combine a VLP encoder and T5 encoder in parallel (i.e., fusing visual features when decoding). 
For knowledge exploration, their main focus is how to use the regional feature to retrieve Wikipedia and GPT-3, while we aim to explore and use more comprehensive types of knowledge, such as prompting OFA to obtain implicit multimodal knowledge. 


\subsection{ VLP Models and PLMs}\label{moreVLP}
Transformer-based PLMs \cite{devlin2018bert,liu2019roberta,raffel2020exploring} have achieved remarkable success in NLP, with the help of large-scale textual pre-training corpus, such as Wikipedia (2,500M words) and BookCorpus (800M words). Recently, VLP models \cite{li2019visualbert,tan2019lxmert,lu2019vilbert,chen2019uniter,guo2021bilinear,jiang2020defense,li2020unimo,yu2019deep,singh2019towards} have also made significant progress in various multimodal downstream tasks \citep{krishna2017visual,hudson2019gqa,johnson2017clevr,tapaswi2016movieqa,si2022language}. Compared to PLMs, they are considered to contain less knowledge due to the smaller size of their pre-training datasets, such as Visual Genome (0.01M images and 2M image-text pairs). 

We believe that models initialized with PLMs \cite{garderes2020conceptbert,Marino_2021_CVPR,gao2022transform} (e.g., BERT \citep{devlin2018bert}, T5 \citep{raffel2020exploring}) and VLP models \cite{wu2022multi,ding2022mukea,shevchenko2021reasoning} (e.g., LXMERT \citep{tan2019lxmert}) introduced implicit text knowledge and implicit multimodal knowledge, respectively, which can further enhance model performance as validated by the results in the middle two rows of Table \ref{main}. 

\subsection{ LLMs and Super Large-scale VLP Models}\label{moreLLM}
Recently, the super large-scale language model (LLM) GPT-3 has also been adopted as a knowledge source for OK-VQA. Unlike normal PLMs, GPT-3 is mainly used in a prompt-learning manner without any further fine-tuning. 
Similarly, the very recent VLP model OFA has attracted researchers' attention due to its excellent zero-shot capability for different prompts. To the best of our knowledge, the proposed method is the first to prompt OFA to obtain its implicit multimodal knowledge. 

Inspired by the success of LLMs in NLP, super large-scale visual-language pre-trained models, such as Flamingo \citep{alayrac2022flamingo} and very recent PaLI \citep{chen2022pali}, has also been launched in the multimodal field recently. 
They are pre-trained with a billion-degree multimodal corpus which contains more knowledge than normal VLP models. We also compared our method with these large-scale VLP models in Appendix \ref{moreCompVLP}.

\section{More Implementation Details}\label{moredetail}
We use the OK-VQA dataset of version v1.1\footnote{https://okvqa.allenai.org/download.html} with license CC-BY 4.0\footnote{http://creativecommons.org/licenses/by/4.0/}, containing 9009 training samples and 5046 test samples. Each sample contains an image, a question in English that requires outside knowledge beyond the image to answer correctly, and corresponding ground truth answers annotated by five annotators. 

We use the \textit{lxmert-base-uncased} or \textit{visualbert-vqa} model to initialize the multimodal encoder, and use \textit{T5-large} model to initialize the textual encoder and answer decoder. 
We adopt the \textit{OFA-huge-VQA} version\footnote{All the T5, LXMERT, visualBERT and OFA models are released by huggingface \citep{wolf2020transformers}.} of OFA that is fine-tuned with VQAv2. 
For the multimodal encoder, all the questions are trimmed to the same length of 16 with the tokenizer of BERT, and we use pre-trained Faster R-CNN to extract a set of fixed 36 objects with 2048-dimensional features from each image. 
For the textual encoder, we use the tokenizer of T5 to segment all the input, i.e., ($Q_i, V_i, M_i, (T_i,) p_{i,k}$) into the token sequence with a fixed length of 250 when the number of Wikipedia passages is less than 75.
Note that, to reduce GPU memory usage, when the number of Wikipedia passages is 75, we remove the stop words in Wikipedia passages and set the token sequence length as 200. 
The adopted DPR \citep{karpukhin2020dense} model is pre-trained on multi-question answering datasets \citep{kwiatkowski2019natural, joshi2017triviaqa,berant2013semantic,baudivs2015modeling}. 
The AdamW \citep{loshchilov2017decoupled} optimizer is adopted with a learning rate of 1e-5 for the multimodal encoder and 1e-4 for the textual encoder and the answer decoder, using the linear schedule with warmup. 
We pre-train and finetune the models for 3 epochs with batch sizes of 12 and 8 on A100-80GB, respectively. 
We set the number of Wikipedia passages to 75 when our method combines GPT-3, otherwise 50. 
Following \citep{gao2022transform,lin2022revive}, we apply a normalization process \citep{chen2017reading,lee2019latent} (including whitespace, lowercasing, punctuation and removing articles) for each predictions. Following previous work, all results are abtained by a single run based on same seed. 

\section{More Experimental Results}\label{moreEXP}
\begin{figure}[t]
\centering
\includegraphics[width=0.98\linewidth]{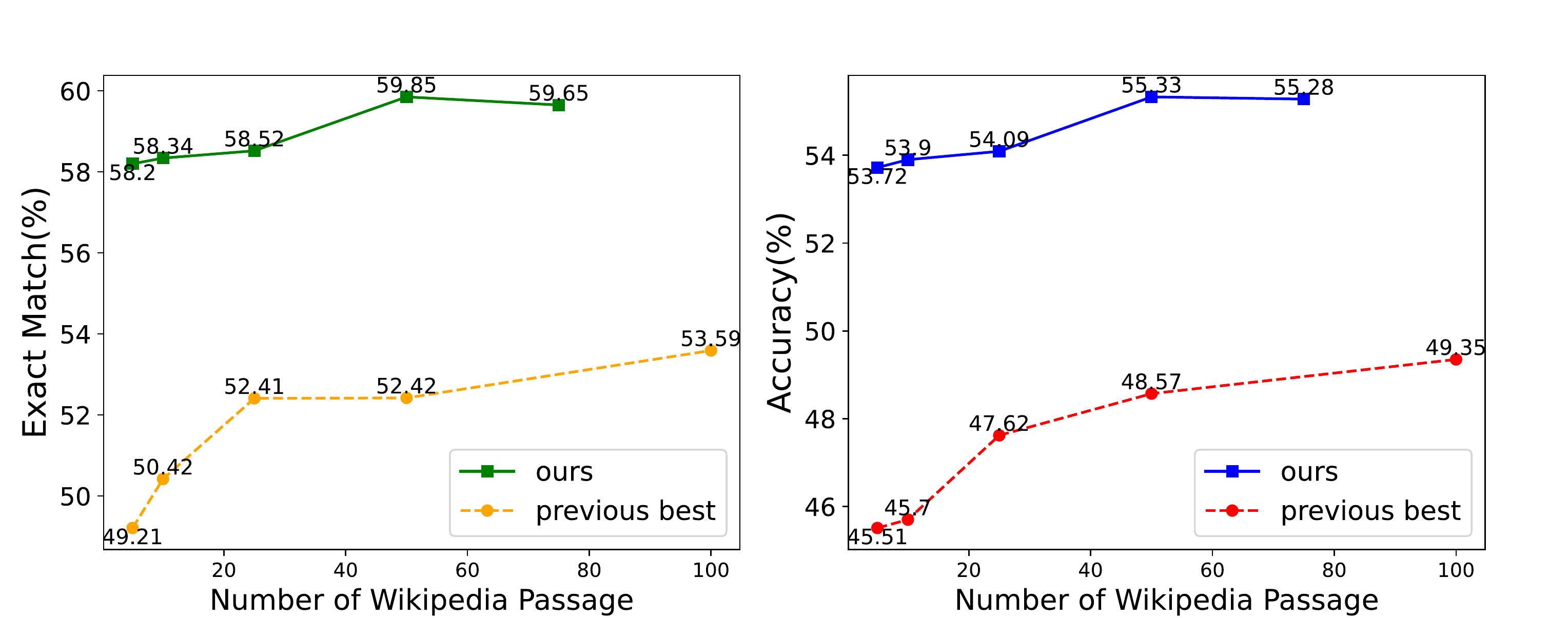} 
\caption{Comparison of EM (left) and accuracy (right) on OK-VQA with varying number of Wikipedia passages. }
\vspace{-0.2cm}
\label{zhexian_copare}
\end{figure}
\subsection{Performance Using Varying Number of Passages}\label{moreWiki}
Figure \ref{zhexian_copare} shows the performance with a varying number of passages, and we find that: 
(1) Our method is consistently superior to the previous-best TRiG no matter with a varying number of Wikipedia passages. With merely 5 passages, the proposed method can perform much better than TRiG with 100 passages, which greatly improves model training and inference speed. 
(2) The performance fluctuation is not as large as before under a different number of Wikipedia passages, which indicates that explicit knowledge in Wikipedia is no longer the only major source of knowledge. 
(3) With the increase in the number of Wikipedia passages, the performance of our model increases first and then decreases. This can be explained by the low recall rate of knowledge when the number of passages is small, while noise is introduced when the number of passages is large.

\subsection{Comparison with Super Large-scale VLP Models}\label{moreCompVLP}

\begin{table}
\centering
\resizebox{0.8\linewidth}{!}{ 
\begin{tabular}{c|c|c|c}
\hline
\textbf{Method} & \textbf{\#Params} & \textbf{\#Pre. Data} & \textbf{Acc}\\
\hline
Flamingo&80B & 2.3B & 57.80 \\
PaLI&3B & 1.6B &52.40\\
PaLI&15B & 1.6B &56.50\\
PaLI&17B & 1.6B &64.50\\ \hline
ours-LXM &0.98B \small(+0.93B)  & 0.44M & 56.49\\ 
ours-ViB &0.88B \small(+0.93B)  & 0.44M & 56.67\\ 



\hline
\end{tabular}
}
\caption{\label{LVLP}
Results comparison with super large-scale visual-language pre-trained models. "\#Pre. Data" represents the size of pre-training data. (+0.93B) represents the parameter quantity of OFA used offline. 
}
\end{table}

Table \ref{LVLP} shows the excellent performance of super large-scale VLP models on OK-VQA. However, they 
 are difficult to deploy 
 due to the huge number of parameters. 
Our method achieved competitive results with these models, using much fewer parameters and only 0.03\% data for pre-training.

\begin{figure}[t]
\centering
\includegraphics[width=0.225\textwidth]{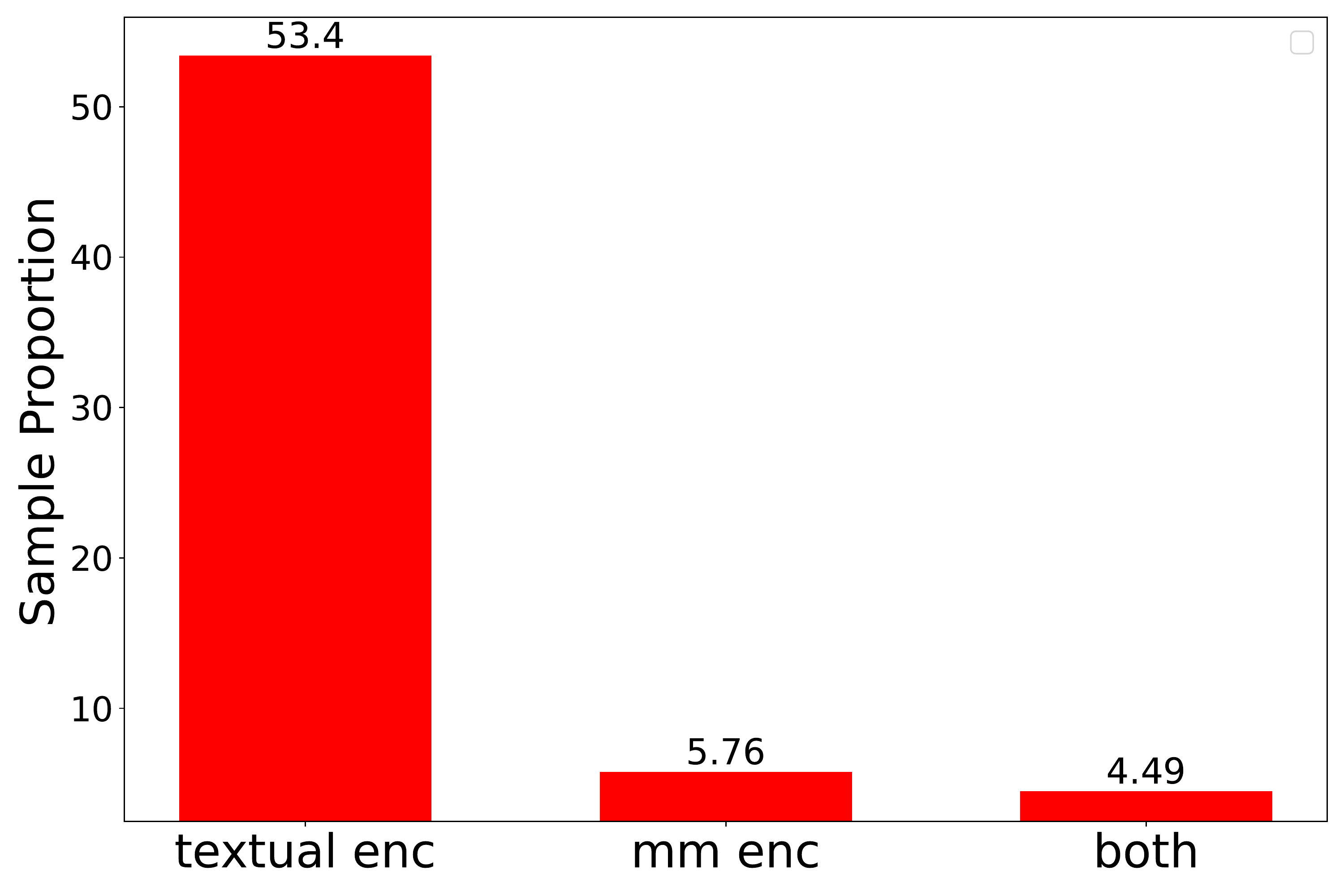} 
\includegraphics[width=0.225\textwidth]{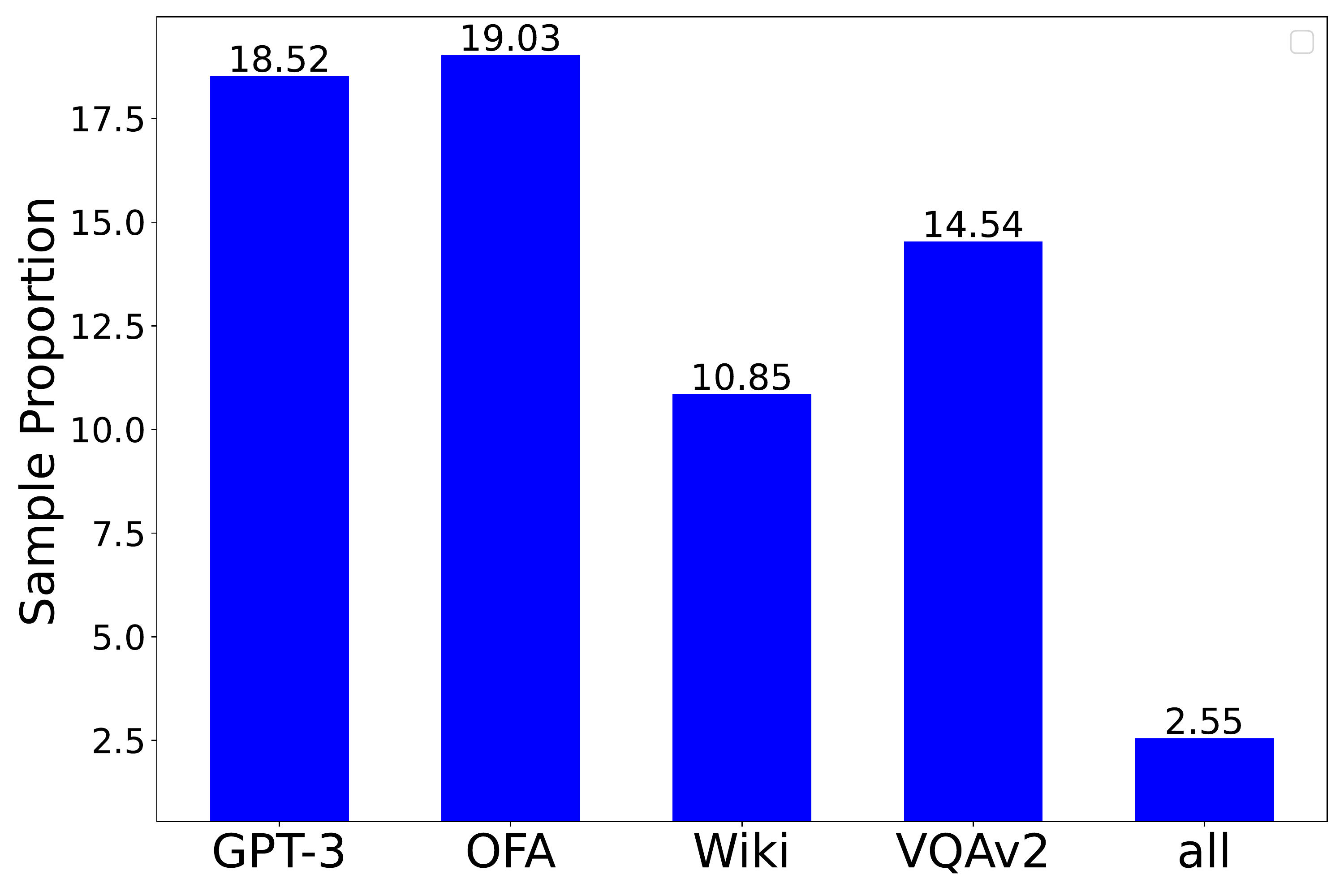} 
\caption{The proportion of the correctly-answered samples that will be answered incorrectly when the certain (left) encoder, (right) knowledge source is removed. } 
\label{zhuzhuang}
\end{figure}

\subsection{ Ablations from a Dependence
Perspective}\label{moreAblation}
As shown in the left part of Figure \ref{zhuzhuang}, we analyze the contribution of the two encoders in our final performance from another perspective. 
53.40\% and 5.76\% of the correctly-answered samples rely on the textual encoder and multimodal encoder, respectively, as they will be answered incorrectly when removing the textual encoder or multimodal encoder. 
Moreover, 4.49\% of samples can only be answered correctly by relying on both encoders at the same time, which indicates that both encoders are indispensable. 

From the right part of Figure \ref{zhuzhuang}, it can be seen that 10.85\%\textasciitilde 19.03\% of correctly answered samples will go wrong if any of the knowledge types are missing. 
This high proportion indicates that all types of knowledge\footnote{Implicit knowledge in T5 and LXMERT will not be discussed here, since they are considered as the parts of the model structure.} are complementary to each other for our method. 
Moreover, 2.25\% of samples can only be answered correctly when all four types of knowledge are available, which proves that more comprehensive knowledge is necessary. 

\section{More Discussion and Qualitative Analysis}\label{moreDis}
\subsection{Conversion Rate from Knowledge to Answers}\label{moreConversion}
\begin{figure}[t]
\centering
\includegraphics[width=0.9\columnwidth]{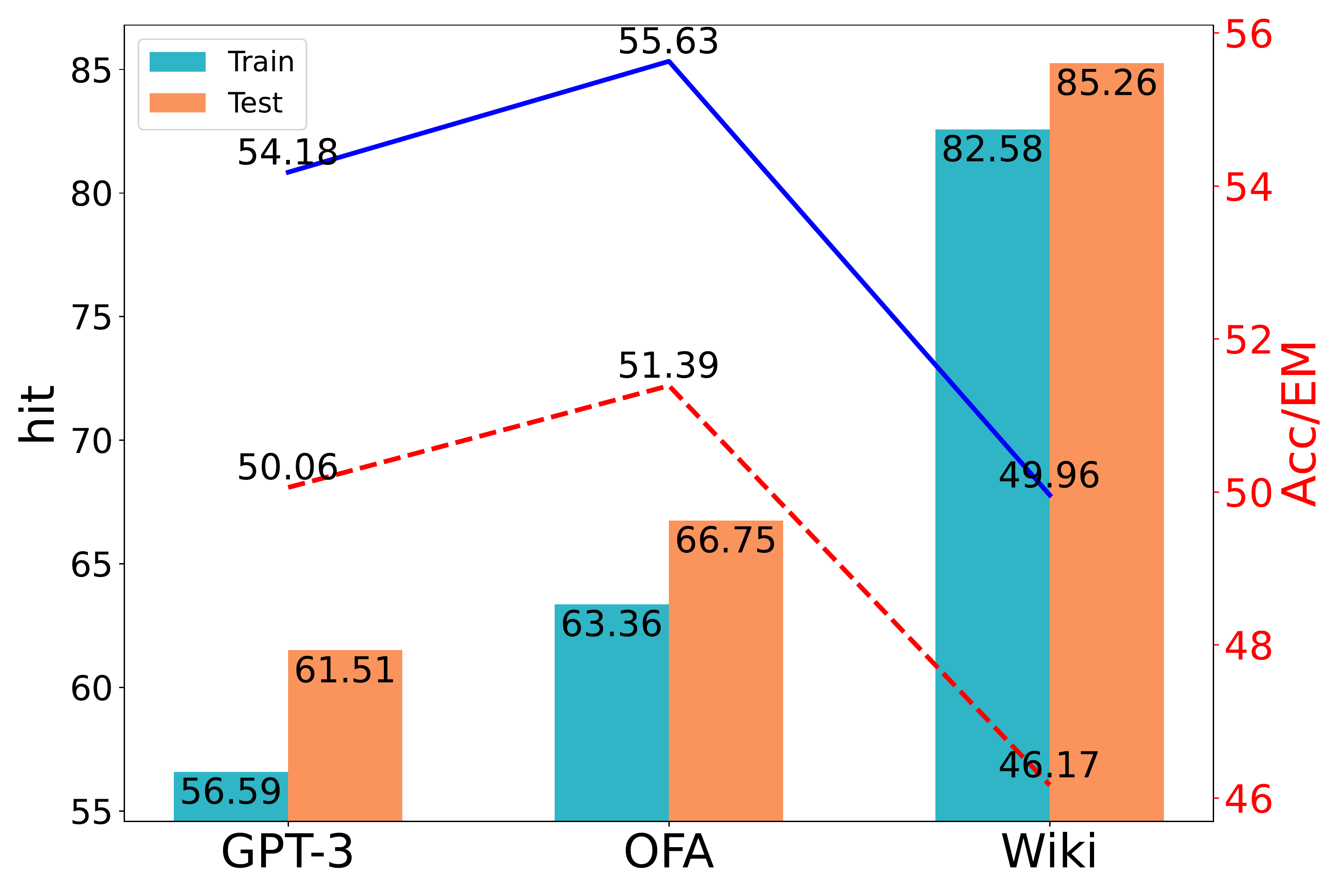} 
\caption{$Hit$ of the knowledge retrieved from GPT-3, OFA, and Wikipedia. The red and blue lines respectively represent the Acc and EM of the models that only bring in one type of knowledge without pre-training. The smaller $hit$ and the greater Acc/EM, the higher the conversion rate of this type of knowledge to the answer.   }
\label{zhuanhua}
\end{figure}


To further explore the extent to which the model makes use of each type of knowledge, we conduct experiments to evaluate the conversion rate of knowledge to the correct answers. 
Note that the explicit multimodal knowledge in VQAv2 is introduced in the manner of pre-training, it is thus difficult to evaluate its $hit$, and will not be discussed here. 

As shown in Figure \ref{zhuanhua}, OFA (0.93B) recalls correct answers for more samples than GPT (175B). This shows that a unifying VLP model is more suitable for retrieving related knowledge in OK-VQA than an LLM. 
Moreover, although the $hit$ of Wikipedia is much higher than that of GPT-3 or OFA, its Acc/EM is lower than the others by a wide margin.
This shows that higher $hit$ does not necessarily lead to higher Acc/EM, and how to further extract answers from the retrieved knowledge will be an impressive direction in future work. 
On the whole, compared with explicit knowledge, implicit knowledge has a higher conversion rate from knowledge to correct answers.

\subsection{OFA vs OFA-vqa}\label{moreOFA} 

\begin{table}
\centering
\resizebox{1\linewidth}{!}{ 
\begin{tabular}{l|c|c|c|c}
\hline

Model & zero-shot Acc & ans $hit$ & ans + evi $hit$ & Acc \\ \hline
OFA & \textbf{33.57} & 45.35 & 49.53 & 52.67 \\ \hline
OFA-vqa & 3.26 & \textbf{57.63} & \textbf{63.36} & \textbf{55.33} \\ \hline

\hline
\end{tabular}
}
\caption{\label{ofa}
Comparison between different versions of OFA.
}
\end{table}
\begin{figure}[t]
\centering
\includegraphics[width=0.95\columnwidth]{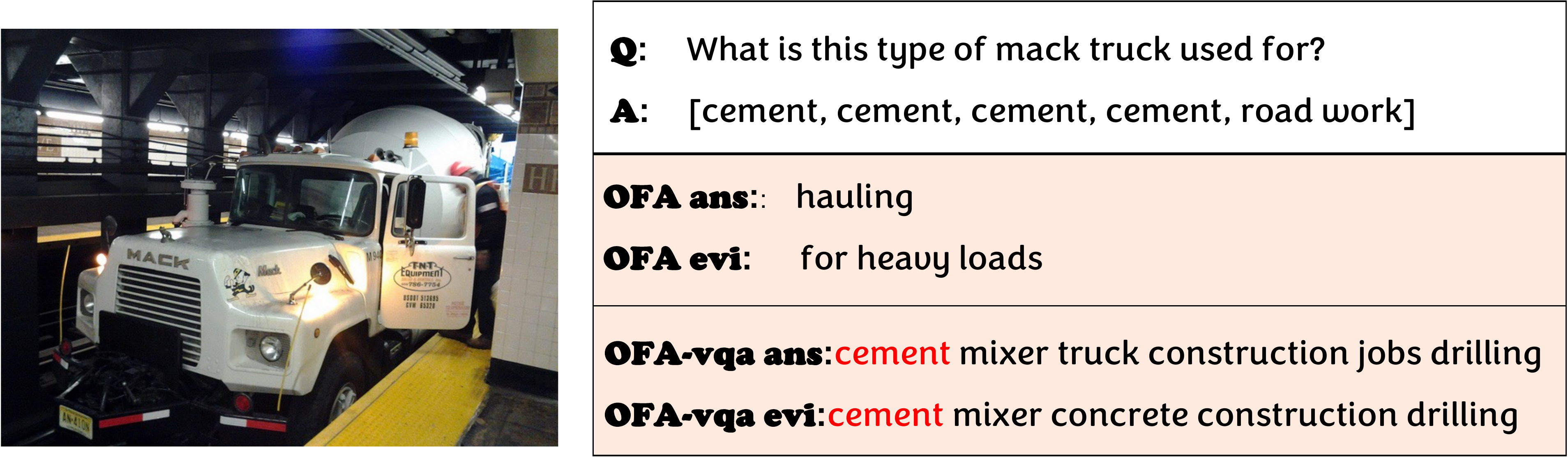} 
\caption{ Comparison of knowledge retrieved from OFA and OFA-vqa. "ans" and "evi" represent the tentative answers and the supporting evidence, respectively. }
\label{ofa-fig}
\end{figure}
OFA releases many versions of models, including VQA-vqa which is fine-tuned on VQAv2 dataset. As shown in Table \ref{ofa}, we compare the performance of the two versions and find that OFA-vqa has improved the $hit$ of knowledge at the expense of the accuracy of its direct testing in OK-VQA and the natural fluency of the language (see Figure \ref{ofa-fig}). In order to introduce more knowledge, we adopted OFA-vqa version and further improved the model performance. Note that due to the dataset bias in VQAv2 (i.e., the answer to about half of the questions is "yes" or "no"), the model always inputs the adhesion of the two items, e.g., "yesno" or "yesyesyes", we thus remove these frequently misspelled words in the output of OFA-vqa.

\subsection{Qualitative Comparison between Ours and Baselines.}\label{morecase}
\begin{figure}[t]
\centering
\includegraphics[width=0.95\columnwidth]{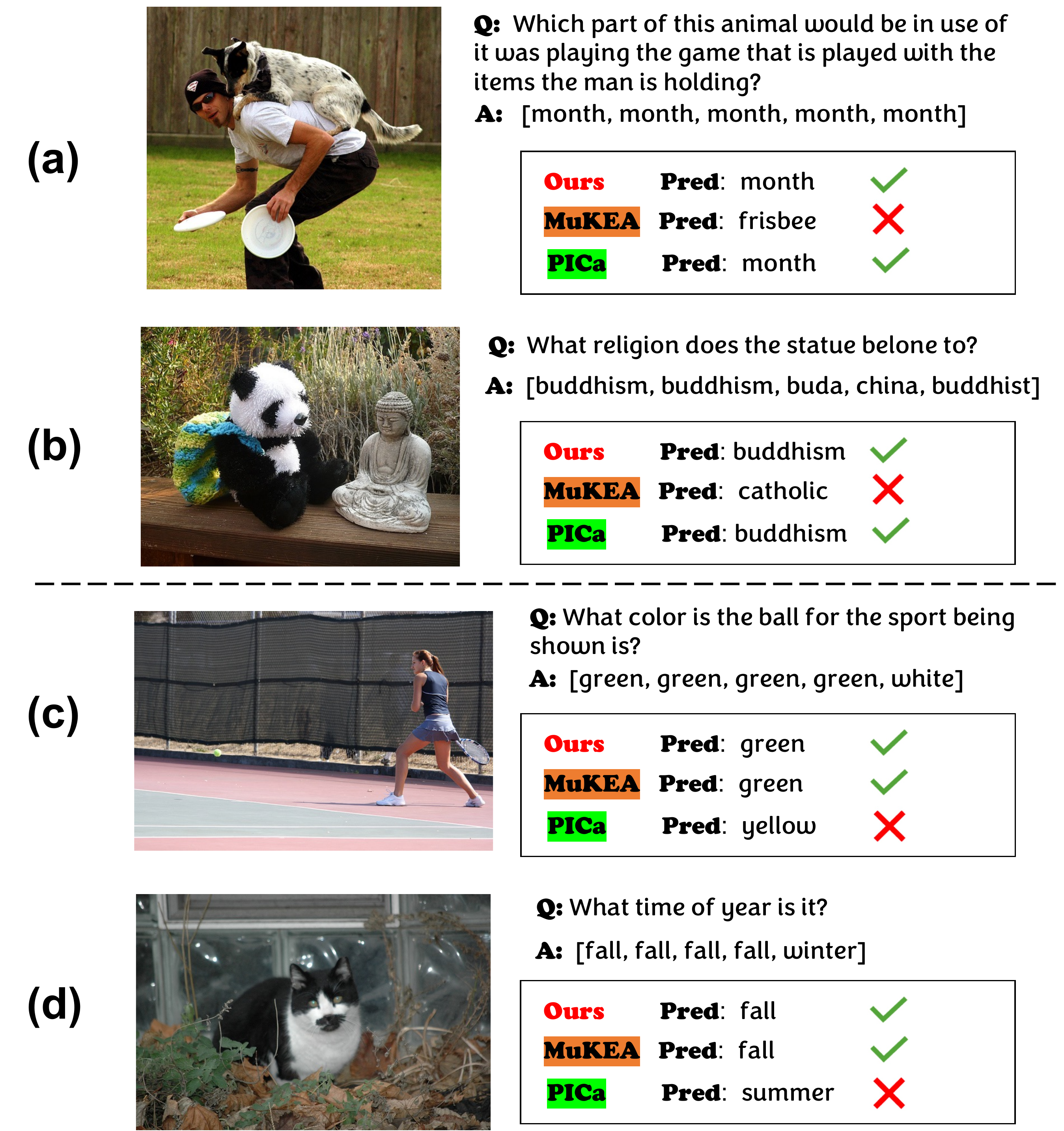} 
\caption{ Qualitative comparison between our method and baselines (MuKEA and PICa). MuKEA is based on the VLP model LXMERT, which explores knowledge in cross-modality space. PICa is based on the LLM model GPT-3, which explores knowledge in natural-language space. \textbf{Pred} denotes the predicted answers. }
\label{caseCompare-fig}
\end{figure}

We qualitatively evaluate the effectiveness of our method in Figure \ref{caseCompare-fig}. The baselines selected here are MuKEA \citep{ding2022mukea} and PICa \citep{yang2022empirical}. 
The former follows the conventional VQA paradigm and predicts answers in a close-set classification manner, while the latter follows the language-centric paradigm and predicts answers in an open-vocabulary generative manner. 

As shown in plot (a),  the question is about "animal parts", while MuKEA's answer is about "sport". Obviously, MuKEA does not correctly understand the meaning of the complex question. This is because the conventional VQA paradigm has poor text comprehension compared to the language-centric paradigm. 
As shown in plot (b), MuKEA mistakenly predicts the answer "buddhism" as "catholicism", since the classification manner is easier driven by the dataset bias \cite{agrawal2016analyzing,manjunatha2019explicit} that "catholicism" appears more frequently in its pre-training and training sets. While PICa generates correct answers for the two examples due to the vast textual knowledge of the natural-language space.

As shown in plots (c) and (d), PICa fails to recognize the "color of the ball" and neglects the "dead leaves" in the image scene, respectively, which are vital to answering the given questions.  While MuKEA correctly predicts the two examples due to the comprehensive visual information in cross-modality space. 

In summary, these examples demonstrate that previous paradigms either lack knowledge or fail to capture visual information. In contrast, our method takes both into account and consistently generates the correct answers for these examples. 
This further reflects the rationality of our motivation to combine both natural-language and cross-modality spaces to achieve a combo of "\textit{thinking and observing}". 

\section{Potential Risks}
A lot of work \cite{agrawal2016analyzing,manjunatha2019explicit} has proved that VQA models are prone to learn the dataset bias. Therefore, our model may be driven by the certain bias in OK-VQA and VQAv2 training sets, such as language bias \citep{agrawal2018don}, multimodal shortcut \cite{dancette2021beyond,si2022language} and harmful stereotypes \cite{hirota2022gender}.
\end{document}